# Efficient Distributed Learning with Sparsity

Jialei Wang*, Mladen Kolar†, Nathan Srebro‡, and Tong Zhang♯

*Department of Computer Science, University of Chicago, IL, USA
†Booth School of Business, University of Chicago, IL, USA
‡Toyota Technological Institute at Chicago, IL, USA
♯Department of Statistics, Rutgers University, NJ, USA

May 25, 2016

**Abstract**

We propose a novel, efficient approach for distributed sparse learning in high-dimensions, where observations are randomly partitioned across machines. Computationally, at each round our method only requires the master machine to solve a shifted $\ell_1$ regularized M-estimation problem, and other workers to compute the gradient on local data. In respect of communication, the proposed approach provably matches the estimation error bound of centralized methods within constant rounds of communications (ignoring logarithmic factors). We conduct extensive experiments on both simulated and real world datasets, and demonstrate encouraging performances on high-dimensional regression and classification tasks.

## 1 Introduction

Many problems in machine learning can be cast as a minimization of the expected loss,

$$\min_{\boldsymbol{\beta}} \mathbb{E}_{\mathbf{X},Y\sim\mathcal{D}}\left[\ell(Y,\langle\mathbf{X},\boldsymbol{\beta}\rangle)\right], \tag{1.1}$$

where $(\mathbf{X},Y)\in\mathcal{X}\times\mathcal{Y}\subseteq\mathbb{R}^p\times\mathcal{Y}$ are drawn from an unknown distribution $\mathcal{D}$ and $\ell(\cdot,\cdot)$ is a convex loss function. Unfortunately the distribution $\mathcal{D}$ is generally not known and the minimizer, $\boldsymbol{\beta}^*$, of (1.1) needs to be approximated on the basis of $N$ observations $\{\mathbf{x}_i,y_i\}_{i=1}^N$ drawn from $\mathcal{D}$. Modern massive data sets, where both $N$ and $p$ are huge, create challenges to classical approaches. One of the challenges, which we address in the paper, is that often observations cannot fit in memory of a single machine, but are rather distributed across $m$ machines. For simplicity, we will assume that $N = nm$ and that $j$-th machine has access to observations $\{\mathbf{x}_{ji},y_{ji}\}_{i=1}^n$. All of our results can be easily generalized for a general $N$. Our particular focus is on the high-dimensional setting where the ambient dimension $p$ is as large, or even larger, as the sample size $n$, but only a subset of variables is predictive, that is, $\mathcal{S} := \text{support}(\boldsymbol{\beta}^*) = \{j\in[p] \mid \beta_j \neq 0\}$ and $s = |\mathcal{S}| \ll p$. Learning a sparse $\boldsymbol{\beta}^*$ in a high-dimensional setting is a well studied statistical problem (Bühlmann and van de Geer, 2011; Hastie et al., 2015), however, it creates unique computational challenges in the distributed setting that we address here.

The main contribution of the paper is a novel algorithm for estimating the minimizer $\boldsymbol{\beta}^*$ of (1.1) in a distributed setting. Our estimator is able to achieve performance of a centralized procedure that has access to all data, while keeping computation and communication costs low. Compared to the existing one-shot estimation approach (Lee et al., 2015b), our method can achieve the same statistical performance faster. If the number of communication rounds is allowed to increase by logarithm on number of total machines, our procedure can keep increasing the statistical performance, until matching the centralized procedure, while keeping the computation time low. Furthermore, these results can be achieved under weaker assumptions on the data generating procedure.

In the paper, we assume the communication occurs in rounds, where in each round, machines exchange messages with the master machine and, between two rounds, the machines only compute based on their local information, which includes local data points and messages received before (Zhang et al., 2013b; Shamir and Srebro, 2014; Arjevani and Shamir, 2015). In a non-distributed setting, efficient estimation procedures need to balance statistical efficiency with computation efficiency (runtime). In a distributed setting, the situation gets more complicated and we need to balance two resources, local runtime and number of rounds of communication, with the statistical error. The local runtime refers to the amount of work each machine needs to do. The number of rounds of communication refers to how often do local machines need to exchange messages with the master machine. We compare our procedure to other algorithm using the aforementioned metrics.

**Notations** We use $[n]$ to denote the set $\{1, \ldots, n\}$. For a vector $a \in \mathbb{R}^n$, we let $\text{support}(a) = \{j : a_j \neq 0\}$ be the support set, $||a||_q$, $q \in [1, \infty)$, the $\ell_q$-norm defined as $||a||_q = (\sum_{i \in [n]} |a_i|^q)^{1/q}$, and $||\mathbf{a}||_\infty = \max_{i \in [n]} |a_i|$. For a matrix $A \in \mathbb{R}^{n_1 \times n_2}$, we use the following element-wise $\ell_\infty$ matrix norms $||A||_\infty = \max_{i \in [n_1], j \in [n_2]} |a_{ij}|$. Denote $\mathbf{I}_n$ as $n \times n$ identity matrix. For two sequences of numbers $\{a_n\}_{n=1}^\infty$ and $\{b_n\}_{n=1}^\infty$, we use $a_n = \mathcal{O}(b_n)$ to denote that $a_n \leqslant C b_n$ for some finite positive constant $C$, and for all $n$ large enough. If $a_n = \mathcal{O}(b_n)$ and $b_n = \mathcal{O}(a_n)$, we use the notation $a_n \asymp b_n$. We also use $a_n \lesssim b_n$ for $a_n = \mathcal{O}(b_n)$ and $a_n \gtrsim b_n$ for $b_n = \mathcal{O}(a_n)$.

### 1.1 Overview of main results

Without loss of generality, let the master machine to be the first machine which has access to local dataset $\{\mathbf{x}_{1i}, y_{1i}\}_{i=1}^n$. We consider the following two baseline estimators of the minimizer $\boldsymbol{\beta}^*$ of (1.1). The Local estimator ignores data available on other machines and computes

$$\widehat{\beta}_{\text{local}} = \arg\min_{\boldsymbol{\beta}} \frac{1}{n} \sum_{i=1}^n \ell(y_{1i}, \langle \mathbf{x}_{1i}, \boldsymbol{\beta} \rangle) + \lambda ||\boldsymbol{\beta}||_1 \tag{1.2}$$

using locally available data. The Local procedure is efficient in both communication and computation, however, the resulting estimation error is large compared to an estimator that uses all of the available data. The other idealized baseline is the centralized estimator that we wish we could compute,

$$\widehat{\beta}_{\text{centralize}} = \arg\min_{\boldsymbol{\beta}} \frac{1}{mn} \sum_{j=1}^m \sum_{i=1}^n \ell(y_{ji}, \langle \mathbf{x}_{ji}, \boldsymbol{\beta} \rangle) + \lambda ||\boldsymbol{\beta}||_1,$$

if storage were available or communication were cheap. The centralized approach achieves the optimal statistical error, however, it is impractical due to expensive communication.

In a related setting, Lee et al. (2015b) studied a one-shot approach to learning $\boldsymbol{\beta}^*$ based on averaging the debiased lasso estimators (Avg-Debias) (Zhang and Zhang, 2013). Under strong assumptions their approach matches the centralized error bound after one round of communication. While an encouraging result, there are limitations to this approach, which our method addresses. In particular, the Avd-Debias method has the following problems:

- The debiasing step in Avg-Debias computationally heavy. In particular, the debiasing step requires solving $\mathcal{O}(p)$ generalized lasso problems, which is computationally prohibitive for high-dimensional problems (Zhang and Zhang, 2013; Javanmard and Montanari, 2014). Our procedure, on the other hand, requires only solving one $\ell_1$ penalized objective in each iteration, which has the same time complexity as computing $\widehat{\beta}_{\text{local}}$ in (1.2). See Section 2 for details.
- Avg-Debias procedure only matches the statistical error of the centralized procedure when the sample size per machine satisfies $n \gtrsim ms^2 \log p$. Our approach improves this sample complexity to $n \gtrsim s^2 \log p$.
- Avg-Debias procedure requires strong conditions on the data generating process. For example, the generalized coherence condition is required [1] on the data matrix to make debiasing work. Such a condition is not needed for consistent high-dimensional estimation in a distributed setting as we show here.

Table 1 summarizes the resources required for the approaches discussed above to solve the distributed sparse linear regression problems.

| Approach | $n \gtrsim ms^2 \log p$ | | $ms^2 \log p \gtrsim n \gtrsim s^2 \log p$ | |
|---|---|---|---|---|
| | Communication | Computation | Communication | Computation |
| Centralize | $n \cdot p$ | $m \cdot \texttt{T}_{\texttt{lasso}}$ | $n \cdot p$ | $m \cdot \texttt{T}_{\texttt{lasso}}$ |
| Avg-Debias | $p$ | $p \cdot \texttt{T}_{\texttt{lasso}}$ | $\times$ | $\times$ |
| This paper | $p$ | $2 \cdot \texttt{T}_{\texttt{lasso}}$ | $\log m \cdot p$ | $\log m \cdot \texttt{T}_{\texttt{lasso}}$ |

Table 1: Comparison of resources required for matching the centralized error bound of various approaches for high-dimensional distributed sparse linear regression problems, where $\texttt{T}_{\texttt{lasso}}$ is the runtime for solving a generalized lasso problem of size $n \times p$.

## 1.2 Related Work

Due to its importance, there is a large body of literature on distributed optimization for modern massive data sets. See for example, (Dekel et al., 2012; Duchi et al., 2012, 2014; Zhang et al., 2013b; Zinkevich et al., 2010; Boyd et al., 2011; Balcan et al., 2012; Yang, 2013; Jaggi et al., 2014; Ma et al., 2015; Shamir and Srebro, 2014; Zhang and Xiao, 2015; Lee et al., 2015a; Arjevani and Shamir, 2015) and reference there in. A popular approach to distributed estimation is averaging estimators formed locally by different machines (Mcdonald et al., 2009; Zinkevich et al., 2010; Zhang et al., 2012; Huang and Huo, 2015). Divide-and-conquer procedures also found applications in statistical inference (Zhao et al., 2014a; Cheng and Shang, 2015; Lu et al., 2016). Shamir and Srebro (2014) and Rosenblatt and Nadler (2014) showed that averaging local estimators at the end will have bad dependence on either condition number or dimensions. Yang (2013), Jaggi et al. (2014), and Ma et al. (2015) studied distributed optimization using stochastic dual coordinate descent. These approaches try to find a good balance between computation and communication. However, their communication complexity bounds have a bad dependence on the condition number, resulting in a procedure which is not better than first-order approaches in terms of communication, such as (proximal) accelerated gradient descent (Nesterov, 1983). Shamir et al. (2014) and Zhang and Xiao (2015) proposed truly communication-efficient distributed optimization algorithms which leveraged the local second-order information, resulting in milder dependence on the condition number, compared to the first-order approaches (Boyd et al., 2011; Shamir and Srebro, 2014; Ma et al., 2015). Lower bounds were studied in (Zhang et al., 2013a; Braverman et al., 2015; Arjevani and Shamir, 2015). However, it is not clear how to extend these approaches with non-smooth objectives, for example, the $\ell_1$ regularized problems.

[1] The generalized coherence requires there exists a matrix $\Theta$, such that $||\widehat{\Sigma}\Theta - I_p||_\infty \lesssim \sqrt{\frac{\log p}{n}}$, where $\widehat{\Sigma}$ is the empirical covariance matrix.

Most of the above mentioned work is focused on estimators that are (asymptotically) linear. Averaging at the end reduces the variance of these estimators, resulting in an estimator that matches the performance of centralized procedure. With $\ell_2$ regularization, Zhang et al. (2013c) studied the averaging the estimations with weaker regularization to avoid the large bias problem, under kernel ridge regression setting. The situation in a high-dimensional setting is not so straightforward, due to the biased induced by the sparsity inducing penalty. Zhao et al. (2014b) illustrated how averaging debiased composite quantile regression estimators can be used for efficient inference in a high-dimensional setting. Averaging debiased high-dimensional estimators was subsequently used in Lee et al. (2015b) for distributed estimation, multi-task learning (Wang et al., 2015), and statistical inference (Battey et al., 2015). Concurrent work of (Jordan et al., 2016) (personal communication) also improved computational efficiency of Lee et al. (2015b) using ideas of Shamir et al. (2014).

## 2 Methodology

In this section, we detail our procedure for estimating the minimizer $\boldsymbol{\beta}^*$ of (1.1) in a distributed setting. Algorithm 1 provides an outline of the steps executed by the worker nodes and the master node.

Let

$$\mathcal{L}_j(\boldsymbol{\beta}) = \frac{1}{n}\sum_{i=1}^{n} \ell(y_{ji}, \langle \mathbf{x}_{ji}, \boldsymbol{\beta}\rangle), \quad j \in [m],$$

be the *empirical* loss at each machine. Our method starts by solving a local $\ell_1$ regularized $M$-estimation program. At iteration $t = 0$, the master machine minimizes the following program

$$\widehat{\boldsymbol{\beta}}_0 = \arg\min \mathcal{L}_1(\boldsymbol{\beta}) + \lambda_0 ||\boldsymbol{\beta}||_1. \tag{2.1}$$

A minimizer $\widehat{\boldsymbol{\beta}}_0$ is broadcasted to all other machines, which use it to compute a gradient of the local loss at $\widehat{\boldsymbol{\beta}}_0$. In particular, each local machine computes $\nabla \mathcal{L}_j(\widehat{\boldsymbol{\beta}}_0)$ and communicates this gradient

**Algorithm 1:** Efficient Distributed Sparse Learning (EDSL).

**Input:** Data $\{\mathbf{x}_{ji}, y_{ji}\}_{j\in[m],i\in[n]}$, loss function $\ell(\cdot,\cdot)$.

**Workers:**

**Initialization:** The master computes local $\ell_1$ regularized loss minimization solution $\widehat{\boldsymbol{\beta}}_0$ as (2.1).

**for** $t = 0, 1, \ldots$ **do**

    **for** $j = 2, 3, \ldots, m$ **do**

        **if** *Receive* $\widehat{\boldsymbol{\beta}}_t$ *from the master* **then**

            Calculate gradient $\nabla\mathcal{L}_j(\widehat{\boldsymbol{\beta}}_t)$ and send to the master.

        **end**

    **end**

    **Master:**

    **if** *Receive* $\{\nabla\mathcal{L}_j(\widehat{\boldsymbol{\beta}}_t)\}_{j=2}^m$ *from all workers* **then**

        Solve the shifted $\ell_1$ regularized problem as (2.2) and obtain solution $\widehat{\boldsymbol{\beta}}_{t+1}$, then broadcast $\widehat{\boldsymbol{\beta}}_{t+1}$ to every worker.

    **end**

**end**

back to the master machine. This constitutes one round of communication. At the iteration $t+1$, the master solve the following shifted $\ell_1$ regularized problem

$$\widehat{\boldsymbol{\beta}}_{t+1} = \arg\min_{\boldsymbol{\beta}} \; \mathcal{L}_1(\boldsymbol{\beta}) + \left\langle \frac{1}{m}\sum_{j\in[m]} \nabla\mathcal{L}_j(\widehat{\boldsymbol{\beta}}_t) - \nabla\mathcal{L}_1(\widehat{\boldsymbol{\beta}}_t), \boldsymbol{\beta} \right\rangle + \lambda_{t+1}||\boldsymbol{\beta}||_1. \tag{2.2}$$

A minimizer $\widehat{\boldsymbol{\beta}}_{t+1}$ is communicated to other machines, which use it to compute the local gradient as before.

Formulation (2.2) is inspired by the proposal in Shamir et al. (2014), where the authors studied distributed optimization for smooth and strongly convex empirical objectives. However, we do not use any averaging scheme, which require additional rounds of communication and, moreover, we add an $\ell_1$ regularization term to ensure consistent estimation in high-dimensions. Different from the distributed first-order optimization approaches, the refined objective (2.2) leverages both global first-order information and local higher-order information. To see this, suppose we set $\lambda_{t+1} = 0$ and that $\mathcal{L}_j(\boldsymbol{\beta})$ is a quadratic objective with invertible hessian. Then we have the following closed form solution for (2.2),

$$\widehat{\boldsymbol{\beta}}_{t+1} = \widehat{\boldsymbol{\beta}}_t - \left(\nabla^2\mathcal{L}_1(\widehat{\boldsymbol{\beta}}_t)\right)^{-1}\left(m^{-1}\sum_{j\in[m]}\nabla\mathcal{L}_j(\widehat{\boldsymbol{\beta}}_t)\right),$$

which is exactly a sub-sampled Newton updating rule. Unfortunately for high-dimensional problems, the Hessian is no longer invertible, and a $\ell_1$ regularization is added to make the solution well behaved. The regularization parameter $\lambda_t$ will be chosen in a way, so that it decreases with the iteration number $t$. As a result we will be able to show that the final estimator performs as well at the centralized solution. We discuss in details how to choose $\lambda_t$ in the following section.

# 3 Theoretical Results

In this section, we present the main theoretical results. The proofs are deferred to Appendix. We start by providing a general estimation error bound on $\widehat{\boldsymbol{\beta}} - \boldsymbol{\beta}^*$, where $\widehat{\boldsymbol{\beta}}$ is obtained using Algorithm 1 and $\boldsymbol{\beta}^*$ is a minimizer of (1.1). Consequences of the main result are illustrated on concrete examples in Section 4. To simplify the presentation, we assume that the domain $\mathcal{X}$ is bounded. Furthermore, our analysis relies on the smoothness conditions of the loss function $\ell(\cdot, \cdot)$.

**Assumption 3.1.** *The loss $\ell(\cdot,\cdot)$ is $L$-smooth with respect to the second argument, that is, $\forall a, b, c \in \mathbb{R}$, we have*

$$\ell'(a,b) - \ell'(a,c) \leqslant L|b-c|.$$

*Furthermore, the loss has a bounded third derivative with respect to the second argument, that is, $\forall a, b \in \mathbb{R}$*

$$|\ell'''(a,b)| \leqslant M.$$

The bounded second and third order derivative for $\ell(\cdot,\cdot)$ is true for popular loss functions used in statistical learning, such as squared loss for regression and logistic loss for classification (Zhang et al., 2013b).

Our analysis also require the notion of restricted strong convexity (Negahban et al., 2012).

**Assumption 3.2.** *The empirical loss function $\mathcal{L}_1$ satisfies the following inequality. For any $\Delta \in \mathcal{C}(S,3)$,*

$$\mathcal{L}_1(\boldsymbol{\beta}^* + \Delta) - \mathcal{L}_1(\boldsymbol{\beta}^*) - \langle \nabla \mathcal{L}_1(\boldsymbol{\beta}^*), \Delta \rangle \geqslant \kappa ||\Delta||_2^2,$$

*where $\mathcal{C}(S,3)$ is a restricted subset in $\mathbb{R}^p$,*

$$\mathcal{C}(S,3) = \{\Delta \in \mathbb{R}^p | ||\Delta_{S^c}||_1 \leqslant 3||\Delta_S||_1\}.$$

The restricted strong convexity is an assumption used for showing consistent estimation in high-dimensions (van de Geer and Bühlmann, 2009; Negahban et al., 2012). It holds with high probability for a wide range of models and designs (see, for example, Negahban et al., 2012; Raskutti et al., 2010; Rudelson and Zhou, 2013, for details).

Our main theoretical result establishes a recursive estimation error bound, which relates the estimation error $||\widehat{\boldsymbol{\beta}}_{t+1} - \boldsymbol{\beta}^*||$ to that of the previous iteration $||\widehat{\boldsymbol{\beta}}_t - \boldsymbol{\beta}^*||_1$.

**Theorem 3.3.** *Suppose Assumption 3.1 and 3.2 holds. Let*

$$\lambda_{t+1} = 2\left|\left|\frac{1}{m}\sum_{j\in[m]} \nabla \mathcal{L}_j(\boldsymbol{\beta}^*)\right|\right|_\infty + 2L\left(\max_{j,i} ||\mathbf{x}_{ji}||_\infty^2\right)||\boldsymbol{\beta}^* - \widehat{\boldsymbol{\beta}}_t||_1\sqrt{\frac{\log(2p/\delta)}{n}} + 2M\left(\max_{j,i}||\mathbf{x}_{ji}||_\infty^3\right)\left(||\widehat{\boldsymbol{\beta}}_t - \boldsymbol{\beta}^*||_1^2\right). \tag{3.1}$$

*Then with probability at least $1-\delta$, we have*

$$||\widehat{\boldsymbol{\beta}}_{t+1} - \boldsymbol{\beta}^*||_1 \leqslant \frac{48s}{\kappa}\left|\left|\frac{1}{m}\sum_{j\in[m]}\nabla \mathcal{L}_j(\boldsymbol{\beta}^*)\right|\right|_\infty + \frac{48sL}{\kappa}\left(\max_{j,i}||\mathbf{x}_{ji}||_\infty^2\right)||\boldsymbol{\beta}^* - \widehat{\boldsymbol{\beta}}_t||_1\sqrt{\frac{\log(2p/\delta)}{n}} + \frac{48sM}{\kappa}\left(\max_{j,i}||\mathbf{x}_{ji}||_\infty^3\right)\left(||\widehat{\boldsymbol{\beta}}_t - \boldsymbol{\beta}^*||_1^2\right),$$

*and*

$$||\widehat{\boldsymbol\beta}_{t+1}-\boldsymbol\beta^*||_2 \leqslant \frac{12\sqrt{s}}{\kappa}\left|\left|\frac{1}{m}\sum_{j\in[m]}\nabla\mathcal{L}_j(\boldsymbol\beta^*)\right|\right|_\infty + \frac{12\sqrt{s}L}{\kappa}\left(\max_{j,i}||\mathbf{x}_{ji}||_\infty^2\right)||\boldsymbol\beta^*-\widehat{\boldsymbol\beta}_t||_1\sqrt{\frac{\log(2p/\delta)}{n}} + \frac{4\sqrt{s}M}{\kappa}\left(\max_{j,i}||\mathbf{x}_{ji}||_\infty^3\right)\left(||\widehat{\boldsymbol\beta}_t-\boldsymbol\beta^*||_1^2\right).$$

Theorem 3.3 upper bounds the estimation error $||\widehat{\boldsymbol\beta}_{t+1}-\boldsymbol\beta^*||_1$ as a function of $||\widehat{\boldsymbol\beta}_t-\boldsymbol\beta^*||_1$. Thus by applying Theorem 3.3 iteratively, we immediately obtain the following estimation error bound which depends on the quality of local $\ell_1$ regularized estimation $||\widehat{\boldsymbol\beta}_0-\boldsymbol\beta^*||_1$.

**Corollary 3.4.** *Suppose the conditions of Theorem 3.3 are satisfied. Furthermore, suppose that for all $t$, we have*

$$M\left(\max_{j,i}||\mathbf{x}_{ji}||_\infty\right)||\widehat{\boldsymbol\beta}_t-\boldsymbol\beta^*||_1 \leqslant L\sqrt{\frac{\log(2p/\delta)}{n}}. \tag{3.2}$$

*Then with probability at least $1-\delta$, we have*

$$||\widehat{\boldsymbol\beta}_{t+1}-\boldsymbol\beta^*||_1 \leqslant (1-a_n)^{-1}(1-a_n^{t+1})\cdot\frac{48s}{\kappa}\cdot\left|\left|\frac{1}{m}\sum_{j\in[m]}\nabla\mathcal{L}_j(\boldsymbol\beta^*)\right|\right|_\infty + a_n^{t+1}||\widehat{\boldsymbol\beta}_0-\boldsymbol\beta^*||_1 \tag{3.3}$$

*and*

$$||\widehat{\boldsymbol\beta}_{t+1}-\boldsymbol\beta^*||_2 \leqslant (1-a_n)^{-1}(1-a_n^{t+1})\cdot\frac{12\sqrt{s}}{\kappa}\cdot\left|\left|\frac{1}{m}\sum_{j\in[m]}\nabla\mathcal{L}_j(\boldsymbol\beta^*)\right|\right|_\infty + a_n^t b_n\cdot||\widehat{\boldsymbol\beta}_0-\boldsymbol\beta^*||_1, \tag{3.4}$$

*where*

$$a_n = \frac{96sL}{\kappa}\left(\max_{j,i}||\mathbf{x}_{ji}||_\infty^2\right)\sqrt{\frac{\log(2p/\delta)}{n}} \quad \text{and} \quad b_n = \frac{24\sqrt{s}L}{\kappa}\left(\max_{j,i}||\mathbf{x}_{ji}||_\infty^2\right)\sqrt{\frac{\log(2p/\delta)}{n}}.$$

Condition in (3.2) always holds for quadratic loss, since $M=0$. For other types of losses, condition in (3.2) may not be true for all $t\geqslant 0$. However, since we want $\widehat{\boldsymbol\beta}_t$ to be competitive to $\widehat{\boldsymbol\beta}_{\text{centralize}}$ where $||\widehat{\boldsymbol\beta}_{\text{centralize}}-\boldsymbol\beta^*||_1 \lesssim s\sqrt{\frac{\log p}{mn}}$, when $m\gtrsim s^2$ the condition will hold for $t$ large enough, leading to local exponential rate of convergence.

## 3.1 Sketch of Proof

We first analyze how the estimation error bound decreases after one round of communication, that is, how $||\widehat{\boldsymbol\beta}_{t+1}-\boldsymbol\beta^*||$ decrease with $||\widehat{\boldsymbol\beta}_t-\boldsymbol\beta^*||$. Define

$$\widetilde{\mathcal{L}}_1(\boldsymbol\beta,\widehat{\boldsymbol\beta}_t) = \mathcal{L}_1(\boldsymbol\beta) + \left\langle \frac{1}{m}\sum_{j\in[m]}\nabla\mathcal{L}_j(\widehat{\boldsymbol\beta}_t) - \nabla\mathcal{L}_1(\widehat{\boldsymbol\beta}_t), \boldsymbol\beta\right\rangle.$$

Then

$$\nabla\widetilde{\mathcal{L}}_1(\boldsymbol\beta,\widehat{\boldsymbol\beta}_t) = \nabla\mathcal{L}_1(\boldsymbol\beta) + \frac{1}{m}\sum_{j\in[m]}\nabla\mathcal{L}_j(\widehat{\boldsymbol\beta}_t) - \nabla\mathcal{L}_1(\widehat{\boldsymbol\beta}_t).$$

The following lemma bounds the $\ell_\infty$ norm of $\nabla\widetilde{\mathcal{L}}_1(\boldsymbol\beta,\widehat{\boldsymbol\beta}_t)$.

**Lemma 3.5.** *With probability at least* $1-\delta$*, we have*

$$\left\|\nabla\widetilde{\mathcal{L}}_1(\boldsymbol{\beta}^*,\widehat{\boldsymbol{\beta}}_t)\right\|_\infty \leqslant \left\|\frac{1}{m}\sum_{j\in[m]}\nabla\mathcal{L}_j(\boldsymbol{\beta}^*)\right\|_\infty + 2L\left(\max_{j,i}||\mathbf{x}_{ji}||_\infty^2\right)||\boldsymbol{\beta}^*-\widehat{\boldsymbol{\beta}}_t||_1\sqrt{\frac{\log(2p/\delta)}{n}}$$
$$+ M\left(\max_{j,i}||\mathbf{x}_{ji}||_\infty^3\right)\left(||\widehat{\boldsymbol{\beta}}_t-\boldsymbol{\beta}^*||_1^2\right).$$

The lemma bounds the magnitude of the gradient of the loss at optimum point $\boldsymbol{\beta}^*$. This will be used to guide our choice of the $\ell_1$ regularization parameter $\lambda_{t+1}$ in (2.2). The following lemma shows that as long as $\lambda_{t+1}$ is large enough, it is guaranteed that $\widehat{\boldsymbol{\beta}}_{t+1}-\boldsymbol{\beta}^*$ is in a restricted cone.

**Lemma 3.6.** *Suppose* $\lambda_{t+1}$ *is chosen as in* (3.1)*. Then with probability at least* $1-\delta$*, we have* $\widehat{\boldsymbol{\beta}}_{t+1}-\boldsymbol{\beta}^*\in\mathcal{C}(S,3)$*.*

Based on the conic condition and restricted strong convexity condition, we can obtain the recursive error bound stated in Theorem 3.3 following the proof strategy as in Negahban et al. (2012).

# 4 Illustrative Examples

In this section we discuss some representative examples of high-dimensional statistical learning problems, which have been extensively studied in recent years. We will use bounds established in Section 3 to obtain guarantees of the proposed algorithm for these problems. For completeness, we first provide the definition of subgaussian norm (Vershynin, 2012).

**Definition 4.1** (Subgaussian norm)**.** *The subgaussian norm* $||X||_{\psi_2}$ *of a subgaussian* $p$*-dimensional random vector* $X$*, is defined as*

$$||X||_{\psi_2} = \sup_{x\in\mathbb{S}^{p-1}}\sup_{q>1} q^{-1/2}(\mathbb{E}|\langle X,x\rangle|^q)^{1/q},$$

*where* $\mathbb{S}^{p-1}$ *is the* $p$*-dimensional unit sphere.*

## 4.1 Sparse Linear Regression

Sparse linear regression is the most widely studied model in high-dimensional statistics (Bühlmann and van de Geer, 2011). In the sparse linear regression setting, data $\{\mathbf{x}_{ji},y_{ji}\}_{i\in[n],j\in[m]}$ are generated according to the model

$$y_{ji} = \langle\mathbf{x}_{ji},\boldsymbol{\beta}^*\rangle+\epsilon_{ji}, \tag{4.1}$$

where $\epsilon_{ji}$ are i.i.d. mean zero subgaussian random variables. For the regression problem, the typical choice of loss function is the squared loss $\ell(y_{ji},\langle\boldsymbol{\beta},\mathbf{x}_{ji}\rangle)=\frac{1}{2}(y_{ji}-\langle\boldsymbol{\beta},\mathbf{x}_{ji}\rangle)^2$, and adding a $\ell_1$ penalty to the empirical loss leads to the lasso estimator (Tibshirani, 1996)

$$\widehat{\boldsymbol{\beta}}_{\text{centralize}} = \arg\min_{\boldsymbol{\beta}}\frac{1}{2mn}\sum_{j\in[m]}\sum_{i\in[n]}(y_{ji}-\langle\boldsymbol{\beta},\mathbf{x}_{ji}\rangle)^2+\lambda||\boldsymbol{\beta}||_1.$$

It is easy to see that the quadratic loss is 1-smooth. Let $\mathcal{L}_j(\boldsymbol{\beta}) = \frac{1}{2n}\sum_{i\in[n]}(y_{ji} - \langle \boldsymbol{\beta}, \mathbf{x}_{ji}\rangle)^2$. When $\mathbf{x}_{ji}$ are randomly drawn from a subgaussian distribution, $\mathcal{L}_1(\boldsymbol{\beta})$ satisfies the restricted strong convexity condition defined in Assumption (3.2) with high probability as long as $n \gtrsim s\log p$ (Rudelson and Zhou, 2013). Moreover, we have the following control on the quantity $\left|\left|\frac{1}{m}\sum_{j\in[m]}\nabla\mathcal{L}_j(\boldsymbol{\beta}^*)\right|\right|_\infty$.

**Lemma 4.2.** *Suppose $||\epsilon_{ji}||_{\phi_2} \leqslant \sigma$ in model 4.1. Then with probability at least $1-\delta$,*

$$\left|\left|\frac{1}{m}\sum_{j\in[m]}\nabla\mathcal{L}_j(\boldsymbol{\beta}^*)\right|\right|_\infty \leqslant \sigma||\mathbf{x}_{ji}||_\infty\sqrt{\frac{\log(p/\delta)}{mn}}.$$

When $\mathbf{x}_{ji}$ are drawn from a mean zero random subgaussian distribution, then $||\mathbf{x}_{ji}||_\infty$ is upper bounded by constant with high probability.

**Lemma 4.3.** *Suppose $||\mathbf{x}_{ji}||_{\phi_2} \leqslant \sigma_X$. Then with probability at least $1-\delta$, we have*

$$\max_{j\in[m], i\in[n]} ||\mathbf{x}_{ji}||_\infty \leqslant \sigma_X\sqrt{\log(mnp/\delta)}.$$

The following $\ell_1$ error bound is standard for lasso with random design, which was established, for example, in (Wainwright, 2009; Meinshausen and Yu, 2009; Bickel et al., 2009)

**Lemma 4.4.** *Under the model 4.1, we have the following estimation error bound for $\widehat{\boldsymbol{\beta}}_0$ holds with probability at least $1-\delta$:*

$$||\widehat{\boldsymbol{\beta}}_0 - \boldsymbol{\beta}^*||_1 \leqslant \frac{s\sigma\sigma_X}{\kappa}\sqrt{\frac{\log(np/\delta)}{n}}.$$

With above analysis for sparse linear regression model with random design, we are ready to present the results for the estimation error bound.

**Corollary 4.5.** *Under sparse linear regression model (4.1) with subgaussian design matrix and noise, and set $\lambda_{t+1}$ as (3.1). Then with probability at least $1-2\delta$, we have the following estimation error bounds for $t \geqslant 0$:*

$$||\widehat{\boldsymbol{\beta}}_{t+1} - \boldsymbol{\beta}^*||_1 \leqslant \frac{1-a_n^{t+1}}{1-a_n}\frac{48s\sigma\sigma_X}{\kappa}\sqrt{\frac{\log(p/\delta)}{mn}} + a_n^{t+1}\frac{s\sigma\sigma_X}{\kappa}\sqrt{\frac{\log(np/\delta)}{n}}, \tag{4.2}$$

$$||\widehat{\boldsymbol{\beta}}_{t+1} - \boldsymbol{\beta}^*||_2 \leqslant \frac{1-a_n^{t+1}}{1-a_n}\frac{12\sqrt{s}\sigma\sigma_X}{\kappa}\sqrt{\frac{\log(p/\delta)}{mn}} + a_n^{t}b_n\frac{s\sigma\sigma_X}{\kappa}\sqrt{\frac{\log(np/\delta)}{n}}, \tag{4.3}$$

*where*

$$a_n = \frac{96s\sigma\sigma_X}{\kappa}\sqrt{\frac{\log(2p/\delta)}{n}} \quad \text{and} \quad b_n = \frac{24\sqrt{s}\sigma\sigma_X}{\kappa}\sqrt{\frac{\log(np/\delta)}{n}}.$$

**Remark 1.** *We can further simplify the bound and look at the scaling with respect to $n, m, s, p$. When $n \gtrsim s^2\log p$, it is easy to see by choosing*

$$\lambda_{t+1} \asymp \sqrt{\frac{\log p}{mn}} + \sqrt{\frac{\log p}{n}}\left(s\sqrt{\frac{\log p}{n}}\right)^{t+1},$$

*the following error bound holds for the proposed algorithm:*

$$||\widehat{\boldsymbol{\beta}}_{t+1} - \boldsymbol{\beta}^*||_1 \lesssim_P s\sqrt{\frac{\log p}{mn}} + \left(s\sqrt{\frac{\log p}{n}}\right)^{t+2},$$

$$||\widehat{\boldsymbol{\beta}}_{t+1} - \boldsymbol{\beta}^*||_2 \lesssim_P \sqrt{\frac{s\log p}{mn}} + \left(\sqrt{\frac{s\log p}{n}}\right)\left(s\sqrt{\frac{\log p}{n}}\right)^{t+1}.$$

*We compare these bounds to the performance of local and centralized lasso (Wainwright, 2009; Meinshausen and Yu, 2009; Bickel et al., 2009). For $\widehat{\boldsymbol{\beta}}_{\text{local}}$, we have*

$$||\widehat{\boldsymbol{\beta}}_{\text{local}} - \boldsymbol{\beta}^*||_1 \lesssim_P s\sqrt{\frac{\log p}{n}} \quad \text{and} \quad ||\widehat{\boldsymbol{\beta}}_{\text{local}} - \boldsymbol{\beta}^*||_2 \lesssim_P \sqrt{\frac{s\log p}{n}}.$$

*For $\widehat{\boldsymbol{\beta}}_{\text{centralize}}$, we have*

$$||\widehat{\boldsymbol{\beta}}_{\text{centralize}} - \boldsymbol{\beta}^*||_1 \lesssim_P s\sqrt{\frac{\log p}{mn}} \quad \text{and} \quad ||\widehat{\boldsymbol{\beta}}_{\text{centralize}} - \boldsymbol{\beta}^*||_2 \lesssim_P \sqrt{\frac{s\log p}{mn}}.$$

*We see that after one round of communications, by choosing*

$$\lambda_1 \asymp \sqrt{\frac{\log p}{mn}} + \frac{s\log p}{n},$$

*we have*

$$||\widehat{\boldsymbol{\beta}}_1 - \boldsymbol{\beta}^*||_1 \lesssim_P s\sqrt{\frac{\log p}{mn}} + \frac{s^2\log p}{n} \quad \text{and} \quad ||\widehat{\boldsymbol{\beta}}_1 - \boldsymbol{\beta}^*||_2 \lesssim_P \sqrt{\frac{s\log p}{mn}} + \frac{s^{3/2}\log p}{n}.$$

*These bounds match the results in Lee et al. (2015b). Furthermore, when $m \lesssim \frac{n}{s^2\log p}$, match the performance for centralized lasso. Moreover, as long as $t \gtrsim \log m$ and $n \gtrsim s^2\log p$, it is easy to check that $\left(s\sqrt{\frac{\log p}{n}}\right)^{t+1} \lesssim s\sqrt{\frac{\log p}{mn}}$. Therefore,*

$$||\widehat{\boldsymbol{\beta}}_{t+1} - \boldsymbol{\beta}^*||_1 \lesssim_P s\sqrt{\frac{\log p}{mn}} \quad \text{and} \quad ||\widehat{\boldsymbol{\beta}}_{t+1} - \boldsymbol{\beta}^*||_2 \lesssim_P \sqrt{\frac{s\log p}{mn}},$$

*which matches the centralized lasso performance without any additional error term.*

## 4.2 Sparse Logistic Regression

Logistic regression is a popular classification model where the binary label $y_{ji} \in \{-1, 1\}$ is drawn according to a Bernoulli distribution:

$$\mathbb{P}(y_{ji} = 1|\mathbf{x}_{ji}) = \frac{\exp(\langle \mathbf{x}_{ji}, \boldsymbol{\beta}^*\rangle)}{\exp(\langle \mathbf{x}_{ji}, \boldsymbol{\beta}^*\rangle) + 1} \quad \text{and} \quad \mathbb{P}(y_{ji} = -1|\mathbf{x}_{ji}) = \frac{1}{\exp(\langle \mathbf{x}_{ji}, \boldsymbol{\beta}^*\rangle) + 1}. \tag{4.4}$$

For logistic model, performing maximum likelihood estimation (MLE) leads to the logistic loss function $\ell(y_{ji}, \langle \boldsymbol{\beta}, \mathbf{x}_{ji}\rangle) = \log(1 + \exp(-y_{ji}\langle \boldsymbol{\beta}, \mathbf{x}_{ji}\rangle))$. For high-dimensional problems, when we add a $\ell_1$ regularization, we obtain the $\ell_1$ regularized logistic regression model (Zhu and Hastie, 2004; Wu et al., 2009):

$$\widehat{\boldsymbol{\beta}}_{\text{centralize}} = \arg\min_{\boldsymbol{\beta}} \frac{1}{mn} \sum_{j\in[m]} \sum_{i\in[n]} \log(1+\exp(-y_{ji}\langle\boldsymbol{\beta},\mathbf{x}_{ji}\rangle)) + \lambda||\boldsymbol{\beta}||_1.$$

The logistic loss is $\frac{1}{4}$-smooth, let $\mathcal{L}_j(\boldsymbol{\beta}) = \frac{1}{n}\sum_{i\in[n]}\log(1+\exp(-\mathbf{y}_{ji}\langle\boldsymbol{\beta},\mathbf{x}_{ji}\rangle))$, (Negahban et al., 2012) showed that if $\mathbf{x}_{ji}$ are drawn from mean zero distribution with sub-Gaussian tails, then $\mathcal{L}_1(\boldsymbol{\beta})$ satisfies the restricted strong condition (3.2). Moreover, we have the following control on the quantity $\left|\left|\frac{1}{m}\sum_{j\in[m]}\nabla\mathcal{L}_j(\boldsymbol{\beta}^*)\right|\right|_\infty$.

**Lemma 4.6.** *Then we have the following upper bound holds in probability at least $1-\delta$:*

$$\left|\left|\frac{1}{m}\sum_{j\in[m]}\nabla\mathcal{L}_j(\boldsymbol{\beta}^*)\right|\right|_\infty \leqslant ||\mathbf{x}_{ji}||_\infty\sqrt{\frac{2\log p}{mn}}.$$

The following $\ell_1$ error bound states the estimation error for logistic regression with $\ell_1$ regularization, which was established, for example, in (van de Geer, 2008; Negahban et al., 2012).

**Lemma 4.7.** *Under the model (4.4), we have the following estimation error bound for $\widehat{\boldsymbol{\beta}}_0$ holds with probability at least $1-\delta$:*

$$||\widehat{\boldsymbol{\beta}}_0 - \boldsymbol{\beta}^*||_1 \leqslant s\sigma_X\sqrt{\frac{2\log(np/\delta)}{n}}.$$

With above analysis for sparse logistic regression model with random design, we are ready to present the results for the estimation error bound which established local exponential convergence.

**Corollary 4.8.** *Under sparse logistic regression model with random design, and set $\lambda_{t+1}$ as (3.1). If the following condition holds for some $T \geqslant 0$:*

$$||\widehat{\boldsymbol{\beta}}_T - \boldsymbol{\beta}^*||_1 \leqslant 4\sqrt{\frac{\log(2p/\delta)}{n}}. \tag{4.5}$$

*Then with probability at least $1-2\delta$, we have the following estimation error bound for all $t \geqslant T$:*

$$||\widehat{\boldsymbol{\beta}}_{t+1} - \boldsymbol{\beta}^*||_1 \leqslant \frac{1-a_n^{t-T+1}}{1-a_n}\frac{96s\sigma\sigma_X}{\kappa}\sqrt{\frac{\log(p/\delta)}{mn}} + 4a_n^{t-T+1}\sqrt{\frac{\log(2p/\delta)}{n}}, \tag{4.6}$$

$$||\widehat{\boldsymbol{\beta}}_{t+1} - \boldsymbol{\beta}^*||_2 \leqslant \frac{1-a_n^{t-T+1}}{1-a_n}\frac{4\sqrt{s}\sigma\sigma_X}{\kappa}\sqrt{\frac{\log(p/\delta)}{mn}} + 4a_n^{t-T}b_n\sqrt{\frac{\log(2p/\delta)}{n}}, \tag{4.7}$$

*where*

$$a_n = \frac{24s\sigma\sigma_X}{\kappa}\sqrt{\frac{\log(2p/\delta)}{n}} \quad \text{and} \quad b_n = \frac{\sqrt{s}\sigma\sigma_X}{\kappa}\sqrt{\frac{\log(np/\delta)}{n}}.$$

### 4.2.1 High-dimensional Generalized Linear Models

The results are readily extendable to other high-dimensional generalized linear models (McCullagh and Nelder, 1989; van de Geer, 2008), where the response variable $y_{ji} \in \mathcal{Y}$ is drawn from the distribution

$$\mathbb{P}(y_{ji}|\mathbf{x}_{ji}) \propto \exp\left(\frac{y_{ji}\langle \mathbf{x}_{ji}, \boldsymbol{\beta}^* \rangle - \Phi(\langle \mathbf{x}_{ji}, \boldsymbol{\beta}^* \rangle)}{A(\sigma)}\right),$$

where $\Phi(\cdot)$ is a link function and $A(\sigma)$ is a scale parameter. Under the random subgaussian design, as long as the loss function has Lipschitz gradient, then the algorithm and corresponding estimation error bound and be applied.

## 4.3 High-dimensional Graphical Models

The results can also be used for the distributed unsupervised learning setting where the task is to learn a sparse graphical structure that represents the conditional independence between variables. Widely studied graphical models are Gaussian graphical models (Meinshausen and Bühlmann, 2006; Yuan and Lin, 2007) for continuous data and Ising graphical models (Ravikumar et al., 2010) for binary observations. As shown in (Meinshausen and Bühlmann, 2006; Ravikumar et al., 2010), these model selection problems can be reduced to solving parallel $\ell_1$ regularized linear regression and logistic regression problems, respectively. Thus the approach presented in this paper can be readily applicable for these tasks.

# 5 Experiments

In this section we present extensive comparisons between various approaches on both simulated and real world datasets. We run the algorithms for both distributed regression and classification problems. The algorithms to be compared are:

- Local: the first machine just solves a related $\ell_1$ regularized problem (lasso or $\ell_1$ regularized logistic regression) with the optimal $\lambda$, and outputs the solution. Obviously this approach is communication free.

- Centralize: the master gathers all data from different machines together, and solves a centralized $\ell_1$ regularized loss minimization problem with the optimal $\lambda$, and outputs the solution. This approach is communication expensive as all data needs to be communicated, but it usually gives us the best estimation and prediction performance.

- Prox GD: the distributed proximal gradient descent is ran on the $\ell_1$ regularized objective, where we initialized the starting point with the first machine's solution.

- Avg-Debias: the method proposed in Lee et al. (2015b), with fine tuned regularization and hard thresholding parameters. This approach only requires one round of communication, where each machine sends a $p$-dimensional vector. However, Avg-Debias is computationally prohibitive because of the debiasing operation.

- EDSL: the proposed efficient distributed sparse learning approach, where the regularization level at each iteration is fine tuned.

$m = 5$ $\quad$ $m = 10$ $\quad$ $m = 20$

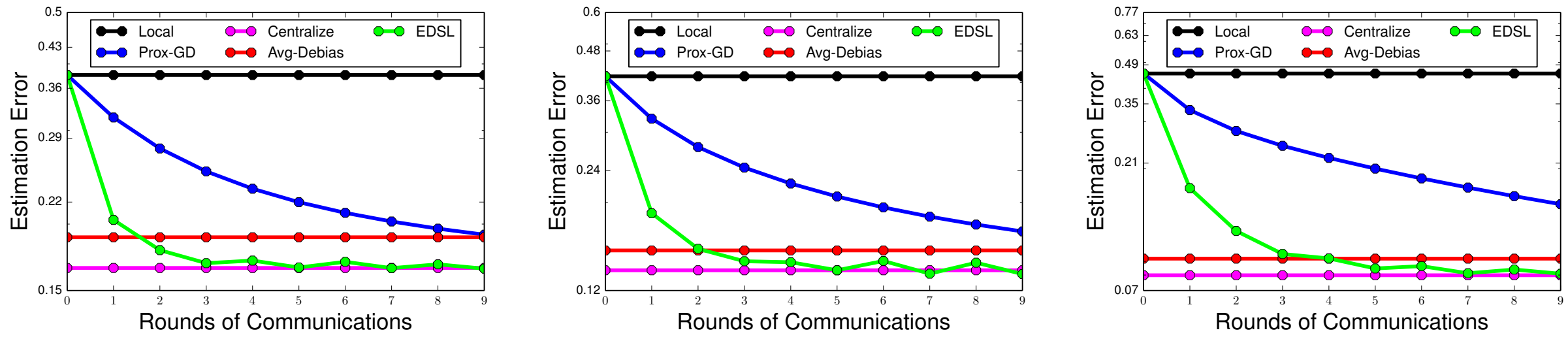

$n = 200, p = 1000, s = 10, \mathbf{X} \sim \mathcal{N}(\mathbf{0}, \mathbf{\Sigma}), \mathbf{\Sigma}_{ij} = 0.5^{|i-j|}$.

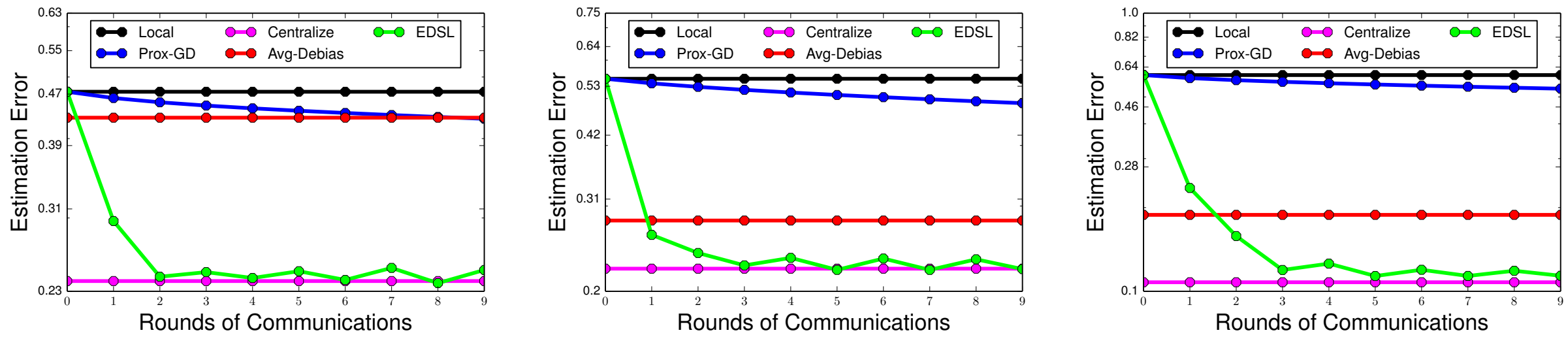

$n = 200, p = 1000, s = 10, \mathbf{X} \sim \mathcal{N}(\mathbf{0}, \mathbf{\Sigma}), \mathbf{\Sigma}_{ij} = 0.5^{|i-j|/5}$.

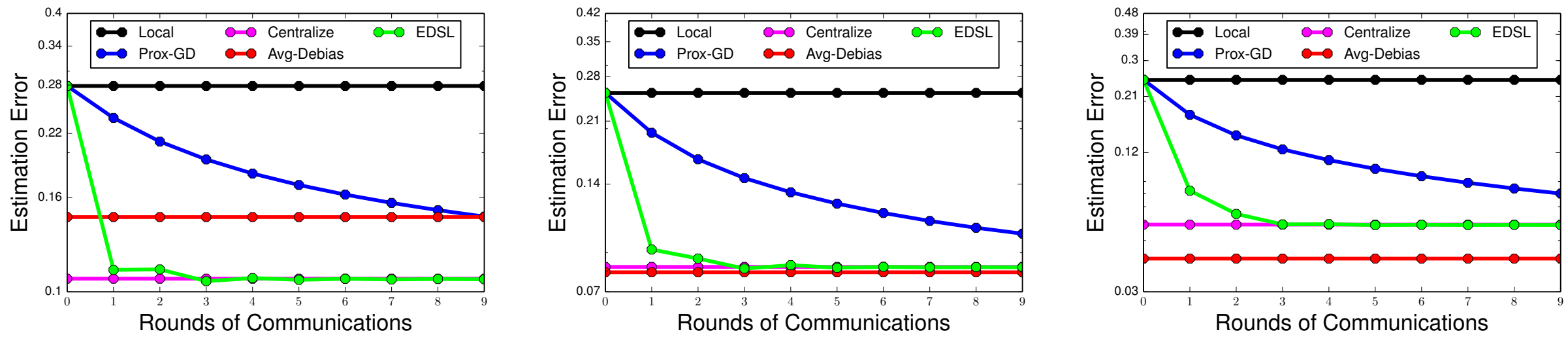

$n = 500, p = 3000, s = 10, \mathbf{X} \sim \mathcal{N}(\mathbf{0}, \mathbf{\Sigma}), \mathbf{\Sigma}_{ij} = 0.5^{|i-j|}$.

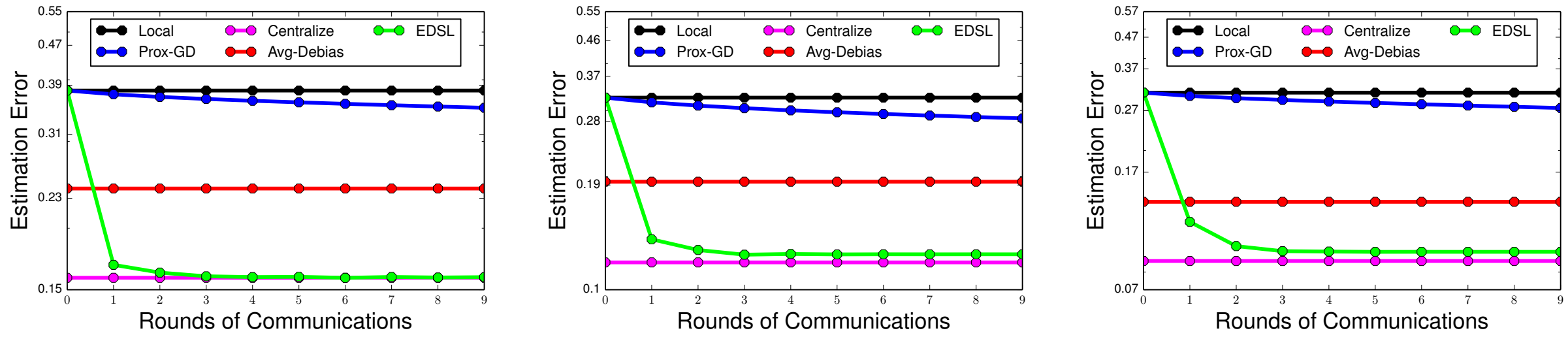

$n = 500, p = 3000, s = 10, \mathbf{X} \sim \mathcal{N}(\mathbf{0}, \mathbf{\Sigma}), \mathbf{\Sigma}_{ij} = 0.5^{|i-j|/5}$.

Figure 1: Comparison of various algorithms for distributed sparse regression, 1st and 3rd row: well-conditioned cases, 2nd and 4th row: ill-conditioned cases.

$m = 5$ $\qquad$ $m = 10$ $\qquad$ $m = 20$

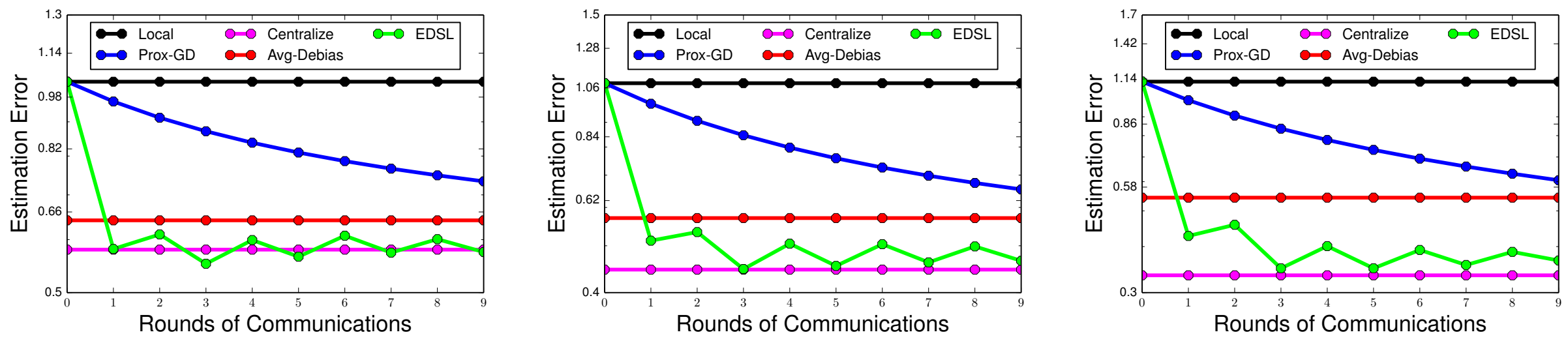

$$n = 500, p = 1000, s = 10, \mathbf{X} \sim \mathcal{N}(\mathbf{0}, \mathbf{\Sigma}), \mathbf{\Sigma}_{ij} = 0.5^{|i-j|}.$$

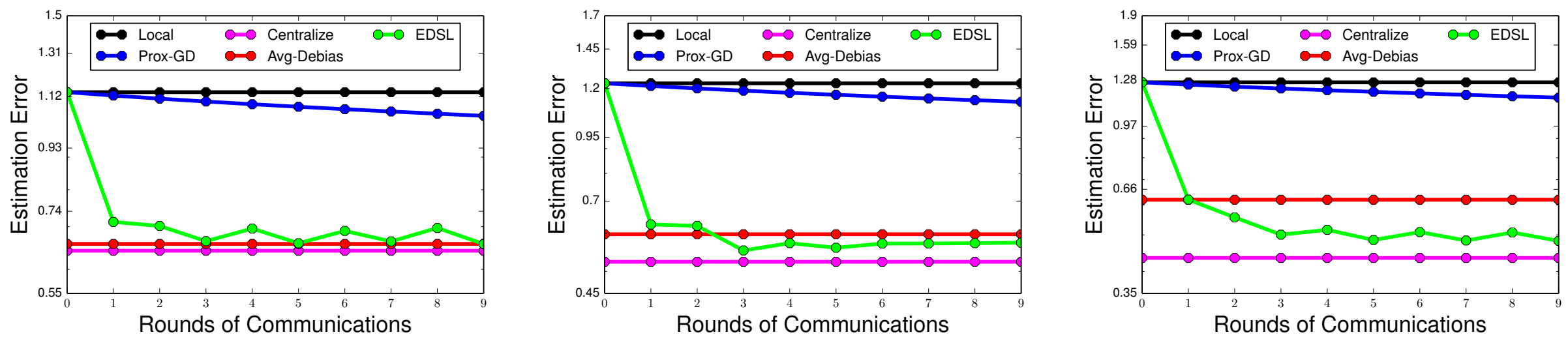

$$n = 500, p = 1000, s = 10, \mathbf{X} \sim \mathcal{N}(\mathbf{0}, \mathbf{\Sigma}), \mathbf{\Sigma}_{ij} = 0.5^{|i-j|/5}.$$

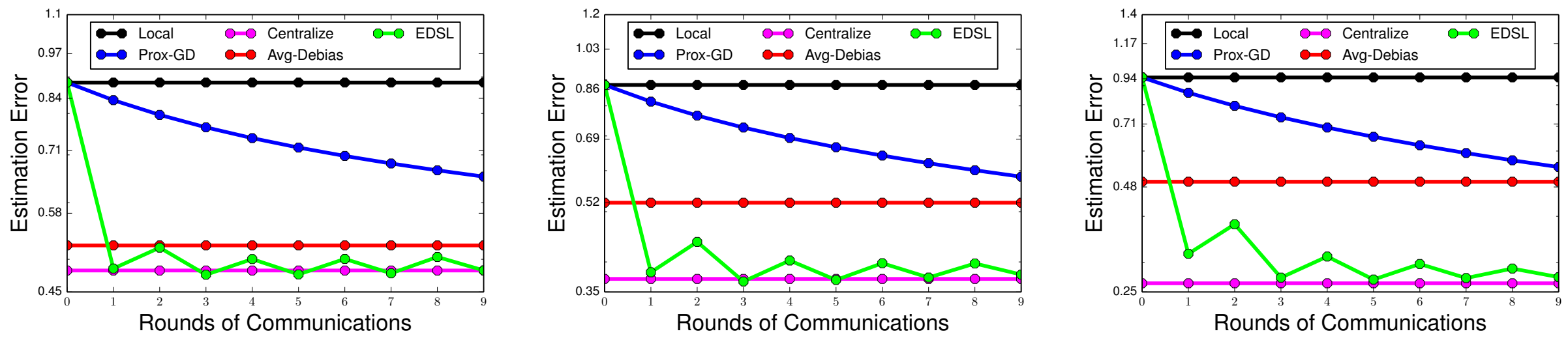

$$n = 1000, p = 3000, s = 10, \mathbf{X} \sim \mathcal{N}(\mathbf{0}, \mathbf{\Sigma}), \mathbf{\Sigma}_{ij} = 0.5^{|i-j|}.$$

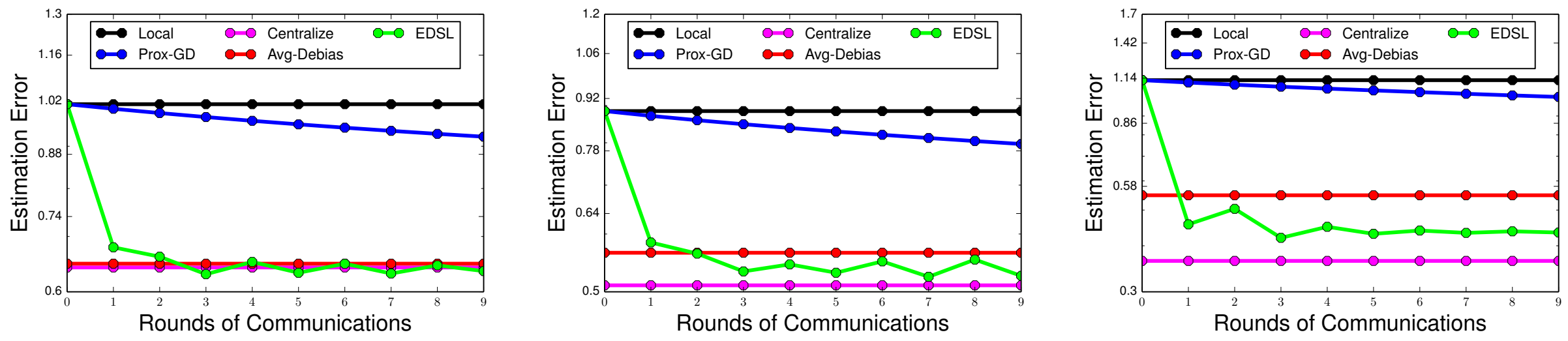

$$n = 1000, p = 3000, s = 10, \mathbf{X} \sim \mathcal{N}(\mathbf{0}, \mathbf{\Sigma}), \mathbf{\Sigma}_{ij} = 0.5^{|i-j|/5}.$$

Figure 2: Comparison of various algorithms for distributed sparse classification (logistic regression), 1st and 3rd row: well-conditioned cases, 2nd and 4th row: ill-conditioned cases.

### 5.1 Simulations

We first examine the algorithms on simulated data. We generate $\{\mathbf{x}_{ji}\}_{j\in[m],i\in[n]}$ from multivariate normal distribution with mean zero vector, and covariance matrix $\boldsymbol{\Sigma}$, which controls the condition number of the problem. We will varying $\Sigma$ to see how it affects the performance of various methods. We set $\Sigma_{ij} = 0.5^{|i-j|}$ for the well-conditioned setting, and $\Sigma_{ij} = 0.5^{|i-j|/5}$ for the ill-conditioned setting. The response variable $\{y_{ji}\}_{j\in[m],i\in[n]}$ are drawn from (4.1) and (4.4) for regression and classification problems, respectively. For regression problems, the noise $\epsilon_{ji}$ is sampled from a standard normal distribution. The true model $\boldsymbol{\beta}^*$ is set to be $s$-sparse, where the first $s$-entries are sampled i.i.d. from a uniform distribution in $[0, 1]$, and the other entries are set to zero.

We run experiments with various $(n, p, m, s)$ settings[2], and plot how the estimation error $||\widehat{\boldsymbol{\beta}}_t - \boldsymbol{\beta}^*||_2$ varies for Prox GD and the proposed EDSL algorithm with rounds of communications. We also plot the estimation error of Local, Avg-Debias, and Centralize as a horizontal line, where the communications are fixed for these algorithms[3]. Figure 1 and 2 summarize the results, where the plots are averaged across 10 independent trials. We have the following observations:

- The Avg-Debias approach obtained much better estimation error than Local after one round of communication, and sometimes performed quite close to Centralize. However, in most cases there is still a gap compared with Centralize, especially when the problem is not well-conditioned, or the number of machines $m$ is large.

- When the problem is well conditioned ($\Sigma_{ij} = 0.5^{|i-j|}$ case), Prox GD converges reasonably fast, however it becomes very slow when the condition number becomes bad ($\Sigma_{ij} = 0.5^{|i-j|/5}$ case). We expect similar phenomenon will to for other first-order distributed optimization algorithms, such as accelerated proximal gradient or ADMM.

- As theory suggested, EDSL obtains a solution that is competitive with Avg-Debias after one round of communication, The estimation error decreases to be truly competitive with the Centralize within very few rounds of communications (typically less than 5, where the theory suggested EDSL will match the performance of Centralize within $\mathcal{O}(\log m)$ rounds of communications).

Above experiments validate the theoretical analysis that when the additional error term in Avg-Debias is relatively large, one round of communication is not sufficient to match the performance of centralized procedure. However, EDSL could match the performance of Avg-Debias also with one round of communication, and further improves the estimation quality by exponentially reducing the additional error until matching the centralized lasso performance, within a few rounds of communications. Thus the proposed EDSL improves the Avg-Debias approach both computationally and statistically.

[2] $n$: sample size per machine, $p$: problem dimension, $m$: number of machines, $s$: true support size.

[3] these algorithms have zero, one-shot and full communications, respectively.

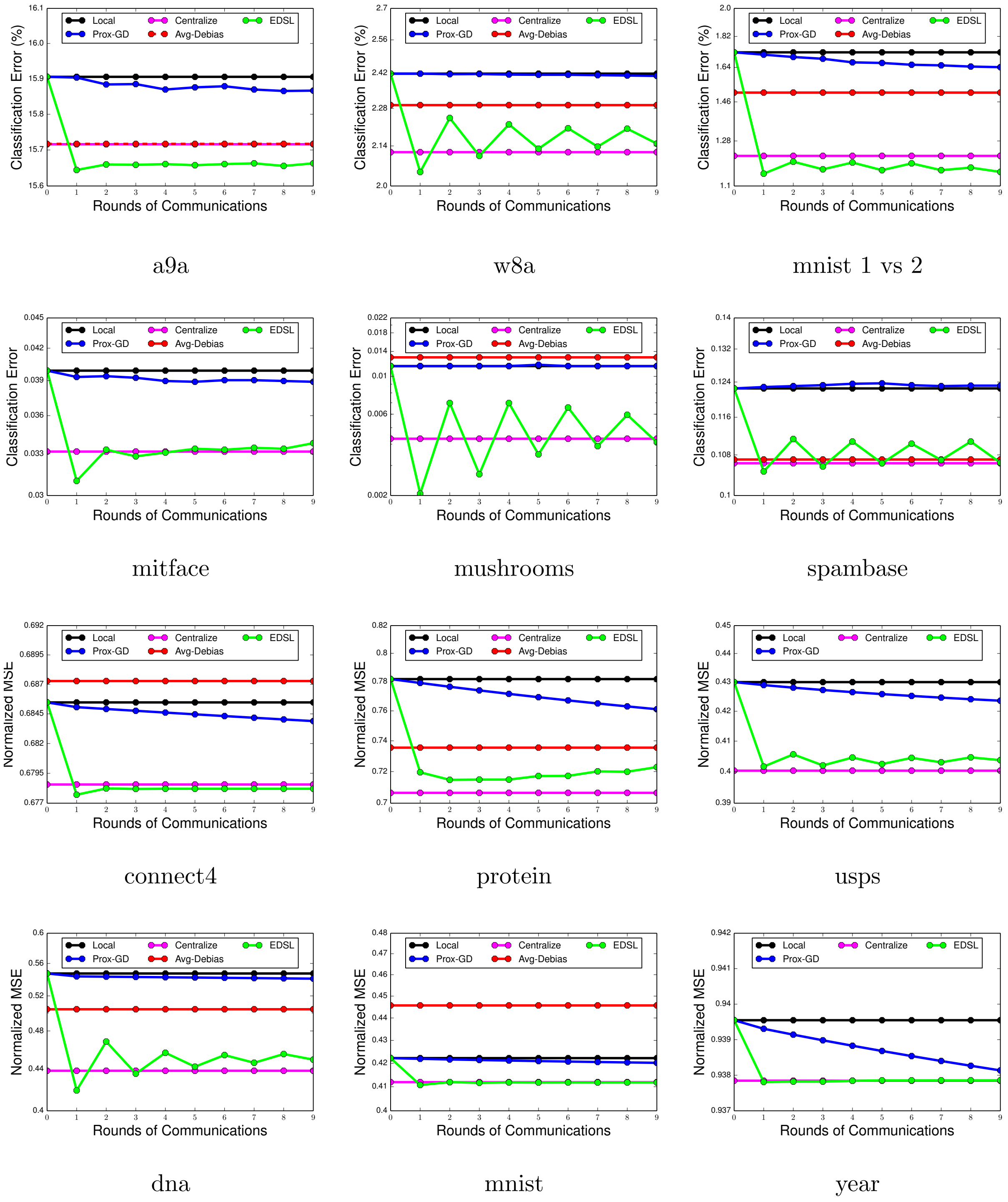

Figure 3: Comparison of various approaches for distributed sparse regression and classification on real world datasets. (Avg-Debias is omitted when it is significantly worse than others.)

Table 2: List of real-world datasets used in the experiments.

| Name | #Instances | #Features | Task |
|---|---|---|---|
| a9a | 48,842 | 123 | Classification |
| connect-4 | 67,557 | 127 | Regression |
| dna | 2,000 | 181 | Regression |
| mitface | 6,977 | 362 | Classification |
| mnist 1 vs 2 | 14,867 | 785 | Classification |
| mnist | 60,000 | 785 | Regression |
| mushrooms | 8,124 | 113 | Classification |
| protein | 17,766 | 358 | Regression |
| spambase | 4,601 | 57 | Classification |
| usps | 7,291 | 257 | Regression |
| w8a | 64,700 | 301 | Classification |
| year | 51,630 | 91 | Regression |

### 5.2 Real-world Data Evaluation

In this section, we compare the distributed sparse learning algorithms on several real world datasets, which are publicly available from the LIBSVM website[4] and UCI Machine Learning Repository[5]. The statistics of these datasets are summarized in Table 2, where some of the multi-class classification datasets are adopted under the regression setting with squared losses. For all of the data, we use 60% of them for training, and 20% as held-out validation set for tuning the parameters, and the remaining 20% for testing. We test 10 randomly partitions of the training, validation and testing sets and report the averaged performance on testing datasets. For regression tasks, the evaluation metric is the normalized Mean Squared Error (normalized MSE), for classification tasks we report the classification error. We randomly partition the data on $m = 10$ machines and run various algorithms tested in the simulation. The results are plotted in Figure 3 where for some datasets the performance of Avg-Debias is significantly worse than others (mostly because the debiasing step fails), thus we omit these plots. We make the following observations based on Figure 3:

- Since there is no well-specified model on these datasets, the curves behave quite differently on different datasets. But overall there is a large gap between the Local solution and Centralized procedure, where the later uses 10 times more data.
- Avg-Debias often fails on these real datasets, and performs much worse than in simulations. The main reason might be becasue when the assumptions such as well-specified model or generalized coherence condition fail, Avg-Debias can totally fail and produce solution even much worse than the local.
- Prox GD approach still converges quite slow in most of the cases.

[4]https://www.csie.ntu.edu.tw/ cjlin/libsvmtools/datasets/

[5]http://archive.ics.uci.edu/ml/

- The proposed EDSL are quite robust on real world datasets, and can output a solution which is highly competitive with the centralized model within a few rounds of communications.
- There exits a "zig-zag" behavior for EDSL approach on some datasets, for example, on mushrooms dataset, the predictive performance of EDSL is not that stable.

The experimental results on real world datasets again verified that the proposed EDSL method is an effective method for distributed sparse learning, while maintain efficiency in both computation and communication.

# 6 Conclusion and Discussion

We proposed a novel approach for distributed learning with sparsity, which is efficient in both computation and communication. Our theoretical analysis showed that the proposed method works under weaker conditions than the averaging the debiased estimator. Furthermore, estimation error can be improved over a few rounds of computation as the additional error term exponentially decreases, until matching the centralized procedure. Extensive experiments on both simulated and real-world data demonstrate that the proposed method improves the performance over one shot averaging approach within just a few rounds of communications.

There might be several ways to improve this work further. As we see in real data experiments, the proposed approach can still perform slightly worse than the centralized approach on certain datasets. It is interesting to explore how to make EDSL provably work under even weaker assumptions. For example, EDSL requires $\mathcal{O}(s^2 \log p)$ samples per machine to match the centralized method in $\mathcal{O}(\log m)$ rounds of communications, however, it is not clear whether the sample size requirement can be improved, while still maintaining low-communication cost. Last but not least, it is interesting to explore the ideas presented to improve the computational cost of communication-efficient distributed multi-task learning with shared support (Wang et al., 2015).

# A Appendix

The appendix contains some theorems and lemmas stated in the main paper.

## A.1 Proof of Lemma 3.5

*Proof.* Based on the definition we know

$$\begin{aligned}\nabla\widetilde{\mathcal{L}}_1(\boldsymbol{\beta}^*,\widehat{\boldsymbol{\beta}}_t) =&\nabla\mathcal{L}_1(\boldsymbol{\beta}^*) + \frac{1}{m}\sum_{j\in[m]}\nabla\mathcal{L}_j(\widehat{\boldsymbol{\beta}}_t) - \nabla\mathcal{L}_1(\widehat{\boldsymbol{\beta}}_t)\\ =&\frac{1}{m}\sum_{j\in[m]}\nabla\mathcal{L}_j(\boldsymbol{\beta}^*) + \nabla\mathcal{L}_1(\boldsymbol{\beta}^*) - \nabla\mathcal{L}_1(\widehat{\boldsymbol{\beta}}_t) - \left(\frac{1}{m}\sum_{j\in[m]}\nabla\mathcal{L}_j(\boldsymbol{\beta}^*) - \frac{1}{m}\sum_{j\in[m]}\nabla\mathcal{L}_j(\widehat{\boldsymbol{\beta}}_t)\right).\end{aligned}$$

For the term $\nabla\mathcal{L}_1(\boldsymbol{\beta}^*) - \nabla\mathcal{L}_1(\widehat{\boldsymbol{\beta}}_t)$, we have

$$\nabla\mathcal{L}_1(\boldsymbol{\beta}^*) - \nabla\mathcal{L}_1(\widehat{\boldsymbol{\beta}}_t) = \frac{1}{n}\sum_{i\in[n]}\mathbf{x}_i(\ell'(y_{1i},\langle\boldsymbol{\beta}^*,\mathbf{x}_{1i}\rangle) - \ell'(y_{1i},\langle\widehat{\boldsymbol{\beta}}_t,\mathbf{x}_{1i}\rangle)).$$

For the term $\frac{1}{m}\sum_{j\in[m]}\nabla\mathcal{L}_j(\boldsymbol{\beta}^*)-\frac{1}{m}\sum_{j\in[m]}\nabla\mathcal{L}_j(\widehat{\boldsymbol{\beta}}_t)$, we have

$$\frac{1}{m}\sum_{j\in[m]}\nabla\mathcal{L}_j(\boldsymbol{\beta}^*)-\frac{1}{m}\sum_{j\in[m]}\nabla\mathcal{L}_j(\widehat{\boldsymbol{\beta}}_t)=\frac{1}{mn}\sum_{j\in[m]}\sum_{i\in[n]}\mathbf{x}_i(\ell'(y_{ji},\langle\boldsymbol{\beta}^*,\mathbf{x}_{ji}\rangle)-\ell'(y_{ji},\langle\widehat{\boldsymbol{\beta}}_t,\mathbf{x}_{ji}\rangle)).$$

Define random vectors $\mathbf{v}_{ji}(\widehat{\boldsymbol{\beta}}_t)\in\mathbb{R}^p$:

$$\mathbf{v}_{ji}(\widehat{\boldsymbol{\beta}}_t)=\mathbf{x}_{ji}(\ell'(y_{ji},\langle\boldsymbol{\beta}^*,\mathbf{x}_{ji}\rangle)-\ell'(y_{ji},\langle\widehat{\boldsymbol{\beta}}_t,\mathbf{x}_{ji}\rangle)).$$

By Taylor series expansion we have

$$\ell'(y_{ji},\langle\widehat{\boldsymbol{\beta}}_t,\mathbf{x}_{ji}\rangle)-\ell'(y_{ji},\langle\boldsymbol{\beta}^*,\mathbf{x}_{ji}\rangle)=\ell''(y_{ji},\langle\boldsymbol{\beta}^*,\mathbf{x}_{ji}\rangle)(\langle\widehat{\boldsymbol{\beta}}_t-\boldsymbol{\beta}^*,\mathbf{x}_{ji}\rangle)+\frac{\ell'''(y_{ji},\mathbf{u}_{ji})}{2}(\langle\widehat{\boldsymbol{\beta}}_t-\boldsymbol{\beta}^*,\mathbf{x}_{ji}\rangle)^2.$$

where $\mathbf{u}_{ji}$ is a number between $\langle\widehat{\boldsymbol{\beta}}_t,\mathbf{x}_{ji}\rangle$ and $\langle\boldsymbol{\beta}^*,\mathbf{x}_{ji}\rangle$. Define $\tau_{ji}=\ell'(y_{ji},\langle\boldsymbol{\beta}^*,\mathbf{x}_{ji}\rangle)$, we have

$$\begin{aligned}\left\|\frac{1}{n}\sum_{i\in[n]}\mathbf{v}_{1i}(\widehat{\boldsymbol{\beta}}_t)-\frac{1}{mn}\sum_j\sum_i\mathbf{v}_{ji}(\widehat{\boldsymbol{\beta}}_t)\right\|_\infty\leqslant&\left\|\frac{1}{n}\sum_i\tau_{1i}\mathbf{x}_{1i}\mathbf{x}_{1i}^T(\widehat{\boldsymbol{\beta}}_t-\boldsymbol{\beta}^*)-\frac{1}{mn}\sum_j\sum_i\tau_{ji}\mathbf{x}_{ji}\mathbf{x}_{ji}^T(\widehat{\boldsymbol{\beta}}_t-\boldsymbol{\beta}^*)\right\|_\infty\\&+M\left(\max_{j,i}||\mathbf{x}_{ji}||_\infty^3\right)\left(||\widehat{\boldsymbol{\beta}}_t-\boldsymbol{\beta}^*||_1^2\right).\end{aligned}$$

We can further upper bound $\left\|\frac{1}{n}\sum_i\tau_{1i}\mathbf{x}_{1i}\mathbf{x}_{1i}^T(\widehat{\boldsymbol{\beta}}_t-\boldsymbol{\beta}^*)-\frac{1}{mn}\sum_j\sum_i\tau_{ji}\mathbf{x}_{ji}\mathbf{x}_{ji}^T(\widehat{\boldsymbol{\beta}}_t-\boldsymbol{\beta}^*)\right\|_\infty$ by

$$\begin{aligned}\left\|\frac{1}{n}\sum_j\tau_{1i}\mathbf{x}_{1i}\mathbf{x}_{1i}^T(\widehat{\boldsymbol{\beta}}_t-\boldsymbol{\beta}^*)-\frac{1}{mn}\sum_j\sum_i\tau_{ji}\mathbf{x}_{ji}\mathbf{x}_{ji}^T(\widehat{\boldsymbol{\beta}}_t-\boldsymbol{\beta}^*)\right\|_\infty\leqslant&\left\|\frac{1}{n}\sum_j\tau_{1i}\mathbf{x}_{1i}\mathbf{x}_{1i}^T-\frac{1}{mn}\sum_j\sum_i\tau_{ji}\mathbf{x}_{ji}\mathbf{x}_{ji}^T\right\|_\infty\\&\cdot||\widehat{\boldsymbol{\beta}}_t-\boldsymbol{\beta}^*||_1.\end{aligned}$$

Also Since

$$\begin{aligned}\left\|\frac{1}{n}\sum_i\tau_{1i}\mathbf{x}_{1i}\mathbf{x}_{1i}^T-\frac{1}{mn}\sum_j\sum_i\tau_{ji}\mathbf{x}_{ji}\mathbf{x}_{ji}^T\right\|_\infty\leqslant&\left\|\frac{1}{n}\sum_{i\in[n]}\tau_{1i}\mathbf{x}_{1i}\mathbf{x}_{1i}^T-\mathbb{E}\left[\tau_{ji}\mathbf{x}_{ji}\mathbf{x}_{ji}^T\right]\right\|_\infty\\&+\left\|\frac{1}{mn}\sum_j\sum_i\tau_{ji}\mathbf{x}_{ji}\mathbf{x}_{ji}^T-\mathbb{E}\left[\tau_{ji}\mathbf{x}_{ji}\mathbf{x}_{ji}^T\right]\right\|_\infty.\end{aligned}$$

Since $|\tau_{ji}|\leqslant L$, by Hoeffding inequality with a union bound over $[p]$, we have with probability at least $1-\delta$,

$$\left\|\frac{1}{n}\sum_{i\in[n]}\tau_{1i}\mathbf{x}_{1i}\mathbf{x}_{1i}^T-\mathbb{E}\left[\tau_{ji}\mathbf{x}_{ji}\mathbf{x}_{ji}^T\right]\right\|_\infty\leqslant L\left(\max_{j,i}||\mathbf{x}_{ji}||_\infty^2\right)\sqrt{\frac{2\log(2p/\delta)}{n}},$$

and

$$\left\|\frac{1}{mn}\sum_j\sum_i\tau_{ji}\mathbf{x}_{ji}\mathbf{x}_{ji}^T-\mathbb{E}\left[\tau_{ji}\mathbf{x}_{ji}\mathbf{x}_{ji}^T\right]\right\|_\infty\leqslant L\left(\max_{j,i}||\mathbf{x}_{ji}||_\infty^2\right)\sqrt{\frac{2\log(2p/\delta)}{mn}}.$$

Combining above, with probability at least $1-\delta$ the following inequality holds:

$$\left|\left|\nabla\widetilde{\mathcal{L}}_1(\boldsymbol{\beta}^*,\widehat{\boldsymbol{\beta}}_t)\right|\right|_\infty \leqslant \left|\left|\frac{1}{m}\sum_{j\in[m]}\nabla\mathcal{L}_j(\boldsymbol{\beta}^*)\right|\right|_\infty + L\left(\max_{j,i}||\mathbf{x}_{ji}||_\infty^2\right)||\boldsymbol{\beta}^*-\widehat{\boldsymbol{\beta}}_t||_1\sqrt{\frac{4\log(2p/\delta)}{n}}$$
$$+ M\left(\max_{j,i}||\mathbf{x}_{ji}||_\infty^3\right)\left(||\widehat{\boldsymbol{\beta}}_t-\boldsymbol{\beta}^*||_1^2\right).$$

□

## A.2 Proof of Lemma 3.6

*Proof.* The proof uses ideas presented in (Negahban et al., 2012). By triangle inequality we have

$$\begin{aligned}||\widehat{\boldsymbol{\beta}}_{t+1}||_1-||\boldsymbol{\beta}^*||_1 =&||\boldsymbol{\beta}^*+(\widehat{\boldsymbol{\beta}}_{t+1}-\boldsymbol{\beta}^*)_{S^c}+(\widehat{\boldsymbol{\beta}}_{t+1}-\boldsymbol{\beta}^*)_S||_1-||\boldsymbol{\beta}^*||_1\\ \geqslant&||\boldsymbol{\beta}^*+(\widehat{\boldsymbol{\beta}}_{t+1}-\boldsymbol{\beta}^*)_{S^c}||_1-||(\widehat{\boldsymbol{\beta}}_{t+1}-\boldsymbol{\beta}^*)_S||_1-||\boldsymbol{\beta}^*||_1\\ =&||(\widehat{\boldsymbol{\beta}}_{t+1}-\boldsymbol{\beta}^*)_{S^c}||_1-||(\widehat{\boldsymbol{\beta}}_{t+1}-\boldsymbol{\beta}^*)_S||_1.\end{aligned}$$

By the optimality of $\widehat{\boldsymbol{\beta}}_{t+1}$ for (2.2), we have

$$\widetilde{\mathcal{L}}_1(\widehat{\boldsymbol{\beta}}_{t+1},\widehat{\boldsymbol{\beta}}_t)+\lambda_{t+1}||\widehat{\boldsymbol{\beta}}_{t+1}||_1-\widetilde{\mathcal{L}}_1(\boldsymbol{\beta}^*,\widehat{\boldsymbol{\beta}}_t)-\lambda_{t+1}||\boldsymbol{\beta}^*||_1\leqslant 0.$$

Thus

$$\widetilde{\mathcal{L}}_1(\widehat{\boldsymbol{\beta}}_{t+1},\widehat{\boldsymbol{\beta}}_t)-\widetilde{\mathcal{L}}_1(\boldsymbol{\beta}^*,\widehat{\boldsymbol{\beta}}_t)+\lambda_{t+1}(||(\widehat{\boldsymbol{\beta}}_{t+1}-\boldsymbol{\beta}^*)_{S^c}||_1-||(\widehat{\boldsymbol{\beta}}_{t+1}-\boldsymbol{\beta}^*)_S||_1)\leqslant 0.$$

By the convexity of $\widetilde{\mathcal{L}}_1(\cdot,\widehat{\boldsymbol{\beta}}_t)$, we further have

$$\widetilde{\mathcal{L}}_1(\widehat{\boldsymbol{\beta}}_{t+1},\widehat{\boldsymbol{\beta}}_t)-\widetilde{\mathcal{L}}_1(\boldsymbol{\beta}^*,\widehat{\boldsymbol{\beta}}_t)\geqslant\langle\nabla\widetilde{\mathcal{L}}_1(\boldsymbol{\beta}^*,\widehat{\boldsymbol{\beta}}_t),\widehat{\boldsymbol{\beta}}_{t+1}-\boldsymbol{\beta}^*\rangle.$$

Thus by Hölder's inequality

$$\begin{aligned}0\geqslant&\langle\nabla\widetilde{\mathcal{L}}_1(\boldsymbol{\beta}^*,\widehat{\boldsymbol{\beta}}_t),\widehat{\boldsymbol{\beta}}_{t+1}-\boldsymbol{\beta}^*\rangle+\lambda_{t+1}(||(\widehat{\boldsymbol{\beta}}_{t+1}-\boldsymbol{\beta}^*)_{S^c}||_1-||(\widehat{\boldsymbol{\beta}}_{t+1}-\boldsymbol{\beta}^*)_S||_1)\\ \geqslant&-||\nabla\widetilde{\mathcal{L}}_1(\boldsymbol{\beta}^*,\widehat{\boldsymbol{\beta}}_t)||_\infty||\widehat{\boldsymbol{\beta}}_{t+1}-\boldsymbol{\beta}^*||_1+\lambda_{t+1}(||(\widehat{\boldsymbol{\beta}}_{t+1}-\boldsymbol{\beta}^*)_{S^c}||_1-||(\widehat{\boldsymbol{\beta}}_{t+1}-\boldsymbol{\beta}^*)_S||_1).\end{aligned}$$

By Lemma 3.5 and (3.1), we know with probability at least $1-\delta$,

$$\begin{aligned}\frac{\lambda_{t+1}}{2}||\widehat{\boldsymbol{\beta}}_{t+1}-\boldsymbol{\beta}^*||_1\geqslant&\left(\left|\left|\frac{1}{m}\sum_{j\in[m]}\nabla\mathcal{L}_j(\boldsymbol{\beta}^*)\right|\right|_\infty+L\left(\max_{j,i}||\mathbf{x}_{ji}||_\infty^2\right)||\boldsymbol{\beta}^*-\boldsymbol{\beta}_t||_1\sqrt{\frac{4\log(2p/\delta)}{n}}\right)\\ &\cdot||\widehat{\boldsymbol{\beta}}_{t+1}-\boldsymbol{\beta}^*||_1+M\left(\max_{j,i}||\mathbf{x}_{ji}||_\infty^3\right)\left(||\widehat{\boldsymbol{\beta}}_t-\boldsymbol{\beta}^*||_1^2\right)||\widehat{\boldsymbol{\beta}}_{t+1}-\boldsymbol{\beta}^*||_1\\ \geqslant&||\nabla\widetilde{\mathcal{L}}_1(\boldsymbol{\beta}^*,\widehat{\boldsymbol{\beta}}_t)||_\infty(||(\widehat{\boldsymbol{\beta}}_{t+1}-\boldsymbol{\beta}^*)_{S^c}||_1-||(\widehat{\boldsymbol{\beta}}_{t+1}-\boldsymbol{\beta}^*)_S||_1)\\ \geqslant&\lambda_{t+1}(||(\widehat{\boldsymbol{\beta}}_{t+1}-\boldsymbol{\beta}^*)_{S^c}||_1-||(\widehat{\boldsymbol{\beta}}_{t+1}-\boldsymbol{\beta}^*)_S||_1-||\widehat{\boldsymbol{\beta}}_{t+1}-\boldsymbol{\beta}^*||_1).\end{aligned}$$

So we obtain

$$\begin{aligned}0\geqslant&||(\widehat{\boldsymbol{\beta}}_{t+1}-\boldsymbol{\beta}^*)_{S^c}||_1-||(\widehat{\boldsymbol{\beta}}_{t+1}-\boldsymbol{\beta}^*)_S||_1-\frac{1}{2}||\widehat{\boldsymbol{\beta}}_{t+1}-\boldsymbol{\beta}^*||_1\\ =&\frac{1}{2}||(\widehat{\boldsymbol{\beta}}_{t+1}-\boldsymbol{\beta}^*)_{S^c}||_1-\frac{3}{2}||(\widehat{\boldsymbol{\beta}}_{t+1}-\boldsymbol{\beta}^*)_S||_1,\end{aligned}$$

which concludes the proof. □

### A.3 Proof of Theorem 3.3

*Proof.* For the term $\widetilde{\mathcal{L}}_1(\widehat{\boldsymbol{\beta}}_{t+1}, \widehat{\boldsymbol{\beta}}_t) - \widetilde{\mathcal{L}}_1(\boldsymbol{\beta}^*, \widehat{\boldsymbol{\beta}}_t)$ we have

$$
\begin{aligned}
\widetilde{\mathcal{L}}_1(\widehat{\boldsymbol{\beta}}_{t+1}, \widehat{\boldsymbol{\beta}}_t) - \widetilde{\mathcal{L}}_1(\boldsymbol{\beta}^*, \widehat{\boldsymbol{\beta}}_t) =& \mathcal{L}_1(\widehat{\boldsymbol{\beta}}_{t+1}) + \left\langle \frac{1}{m} \sum_{j \in [m]} \nabla \mathcal{L}_j(\widehat{\boldsymbol{\beta}}_t) - \nabla \mathcal{L}_1(\widehat{\boldsymbol{\beta}}_t), \widehat{\boldsymbol{\beta}}_{t+1} \right\rangle \\
& - \mathcal{L}_1(\boldsymbol{\beta}^*) - \left\langle \frac{1}{m} \sum_{j \in [m]} \nabla \mathcal{L}_j(\widehat{\boldsymbol{\beta}}_t) - \nabla \mathcal{L}_1(\widehat{\boldsymbol{\beta}}_t), \boldsymbol{\beta}^* \right\rangle \\
\geqslant& \langle \nabla \mathcal{L}_1(\boldsymbol{\beta}^*), \widehat{\boldsymbol{\beta}}_{t+1} - \boldsymbol{\beta}^* \rangle + \kappa ||\widehat{\boldsymbol{\beta}}_{t+1} - \boldsymbol{\beta}^*||_2^2 \\
& + \left\langle \frac{1}{m} \sum_{j \in [m]} \nabla \mathcal{L}_j(\widehat{\boldsymbol{\beta}}_t) - \nabla \mathcal{L}_1(\widehat{\boldsymbol{\beta}}_t), \widehat{\boldsymbol{\beta}}_{t+1} \right\rangle \\
& - \left\langle \frac{1}{m} \sum_{j \in [m]} \nabla \mathcal{L}_j(\widehat{\boldsymbol{\beta}}_t) - \nabla \mathcal{L}_1(\widehat{\boldsymbol{\beta}}_t), \boldsymbol{\beta}^* \right\rangle \\
=& \left\langle \nabla \mathcal{L}_1(\boldsymbol{\beta}^*) + \frac{1}{m} \sum_{j \in [m]} \nabla \mathcal{L}_j(\widehat{\boldsymbol{\beta}}_t) - \nabla \mathcal{L}_1(\widehat{\boldsymbol{\beta}}_t), \widehat{\boldsymbol{\beta}}_{t+1} - \boldsymbol{\beta}^* \right\rangle \\
& + \kappa ||\widehat{\boldsymbol{\beta}}_{t+1} - \boldsymbol{\beta}^*||_2^2 \\
=& \langle \nabla \widetilde{\mathcal{L}}_1(\boldsymbol{\beta}^*, \widehat{\boldsymbol{\beta}}_t), \widehat{\boldsymbol{\beta}}_{t+1} - \boldsymbol{\beta}^* \rangle + \kappa ||\widehat{\boldsymbol{\beta}}_{t+1} - \boldsymbol{\beta}^*||_2^2,
\end{aligned}
$$

where the first inequality we use the restricted strong convexity condition (3.2). Also by the optimality of $\widehat{\boldsymbol{\beta}}_{t+1}$ for (2.2), we have

$$
\widetilde{\mathcal{L}}_1(\widehat{\boldsymbol{\beta}}_{t+1}, \widehat{\boldsymbol{\beta}}_t) - \widetilde{\mathcal{L}}_1(\boldsymbol{\beta}^*, \widehat{\boldsymbol{\beta}}_t) + \lambda_{t+1} ||\widehat{\boldsymbol{\beta}}_{t+1}||_1 - \lambda_{t+1} ||\boldsymbol{\beta}^*||_1 \leqslant 0.
$$

Combining above two inequalities we obtain with probability at least $1 - \delta$:

$$
\begin{aligned}
\lambda_{t+1} ||\boldsymbol{\beta}^*||_1 - \lambda_{t+1} ||\widehat{\boldsymbol{\beta}}_{t+1}||_1 \geqslant& \langle \nabla \widetilde{\mathcal{L}}_1(\boldsymbol{\beta}^*, \widehat{\boldsymbol{\beta}}_t), \widehat{\boldsymbol{\beta}}_{t+1} - \boldsymbol{\beta}^* \rangle + \kappa ||\widehat{\boldsymbol{\beta}}_{t+1} - \boldsymbol{\beta}^*||_2^2 \\
\geqslant& - ||\nabla \widetilde{\mathcal{L}}_1(\boldsymbol{\beta}^*, \widehat{\boldsymbol{\beta}}_t)||_\infty ||\widehat{\boldsymbol{\beta}}_{t+1} - \boldsymbol{\beta}^*||_1 + \kappa ||\widehat{\boldsymbol{\beta}}_{t+1} - \boldsymbol{\beta}^*||_2^2 \\
\geqslant& - \frac{\lambda_{t+1}}{2} ||\widehat{\boldsymbol{\beta}}_{t+1} - \boldsymbol{\beta}^*||_1 + \kappa ||\widehat{\boldsymbol{\beta}}_{t+1} - \boldsymbol{\beta}^*||_2^2.
\end{aligned}
$$

By triangle inequality that $\lambda_{t+1} ||\widehat{\boldsymbol{\beta}}_{t+1} - \boldsymbol{\beta}^*||_1 \geqslant \lambda_{t+1} ||\boldsymbol{\beta}^*||_1 - \lambda_{t+1} ||\widehat{\boldsymbol{\beta}}_{t+1}||_1$, we have

$$
\begin{aligned}
\kappa ||\widehat{\boldsymbol{\beta}}_{t+1} - \boldsymbol{\beta}^*||_2^2 \leqslant& \frac{3\lambda_{t+1}}{2} ||\widehat{\boldsymbol{\beta}}_{t+1} - \boldsymbol{\beta}^*||_1 \\
=& \frac{3\lambda_{t+1}}{2} (||(\widehat{\boldsymbol{\beta}}_{t+1} - \boldsymbol{\beta}^*)_S||_1 + ||(\widehat{\boldsymbol{\beta}}_{t+1} - \boldsymbol{\beta}^*)_{S^c}||_1) \\
\leqslant& \frac{3\lambda_{t+1}}{2} (||(\widehat{\boldsymbol{\beta}}_{t+1} - \boldsymbol{\beta}^*)_S||_1 + 3||(\widehat{\boldsymbol{\beta}}_{t+1} - \boldsymbol{\beta}^*)_S||_1) \\
=& 6\lambda_{t+1} ||(\widehat{\boldsymbol{\beta}}_{t+1} - \boldsymbol{\beta}^*)_S||_1 \\
\leqslant& 6\sqrt{s} \lambda_{t+1} ||(\widehat{\boldsymbol{\beta}}_{t+1} - \boldsymbol{\beta}^*)_S||_2 \\
\leqslant& 6\sqrt{s} \lambda_{t+1} ||\widehat{\boldsymbol{\beta}}_{t+1} - \boldsymbol{\beta}^*||_2.
\end{aligned}
$$

We get

$$||\widehat{\boldsymbol{\beta}}_{t+1} - \boldsymbol{\beta}^*||_2 \leqslant \frac{6\sqrt{s}\lambda_{t+1}}{\kappa}.$$

Substitute $\lambda_{t+1}$ in (3.1) concludes the proof for $\ell_2$ estimation error bound. For $||\widehat{\boldsymbol{\beta}}_{t+1} - \boldsymbol{\beta}^*||_1$, we know

$$\begin{aligned}||\widehat{\boldsymbol{\beta}}_{t+1} - \boldsymbol{\beta}^*||_1 \leqslant & ||(\widehat{\boldsymbol{\beta}}_{t+1} - \boldsymbol{\beta}^*)_S||_1 + ||(\widehat{\boldsymbol{\beta}}_{t+1} - \boldsymbol{\beta}^*)_{S^c}||_1 \\ \leqslant & 4||(\widehat{\boldsymbol{\beta}}_{t+1} - \boldsymbol{\beta}^*)_S||_1 \leqslant 4\sqrt{s}||(\widehat{\boldsymbol{\beta}}_{t+1} - \boldsymbol{\beta}^*)_S||_2 \\ \leqslant & 4\sqrt{s}||\widehat{\boldsymbol{\beta}}_{t+1} - \boldsymbol{\beta}^*||_2 \leqslant \frac{24s\lambda_{t+1}}{\kappa},\end{aligned}$$

which obtains the desired bound. □

### A.4 Proof of Corollary 3.4

*Proof.* The proof proceeds by recursively applying Theorem 3.3 and sum a geometric sequence. For notation simplicity let

$$\begin{aligned}a =& \frac{48s}{\kappa}\left\|\frac{1}{m}\sum_{j\in[m]}\nabla\mathcal{L}_j(\boldsymbol{\beta}^*)\right\|_\infty, \\ b =& \left(\frac{48sL}{\kappa}\left(\max_{j,i}||\mathbf{x}_{ji}||_\infty^2\right)\sqrt{\frac{4\log(2p/\delta)}{n}}\right), \\ c =& \frac{48sM}{\kappa}\left(\max_{j,i}||\mathbf{x}_{ji}||_\infty^3\right).\end{aligned}$$

By Theorem 3.3 we have

$$\begin{aligned}||\widehat{\boldsymbol{\beta}}_{t+1} - \boldsymbol{\beta}^*||_1 \leqslant & a + b||\widehat{\boldsymbol{\beta}}_t - \boldsymbol{\beta}^*||_1 + c||\widehat{\boldsymbol{\beta}}_t - \boldsymbol{\beta}^*||_1^2 \\ \leqslant & a + 2b||\widehat{\boldsymbol{\beta}}_t - \boldsymbol{\beta}^*||_1 \\ \leqslant & a + 2b(a + 2b||\widehat{\boldsymbol{\beta}}_{t-1} - \boldsymbol{\beta}^*||_1) \leqslant \dots \\ \leqslant & a\sum_{k=0}^{t}(2b)^k + (2b)^{t+1}||\widehat{\boldsymbol{\beta}}_0 - \boldsymbol{\beta}^*||_1. \\ = & \frac{a(1-(2b)^{t+1})}{1-2b} + (2b)^{t+1}||\widehat{\boldsymbol{\beta}}_0 - \boldsymbol{\beta}^*||_1, \end{aligned} \tag{A.1}$$

which completes the $\ell_1$ estimation error bound. For $||\widehat{\boldsymbol{\beta}}_{t+1} - \boldsymbol{\beta}^*||_2$, we first use (A.1) to obtain

$$||\widehat{\boldsymbol{\beta}}_t - \boldsymbol{\beta}^*||_1 \leqslant \frac{a(1-(2b)^t)}{1-(2b)} + (2b)^t||\widehat{\boldsymbol{\beta}}_0 - \boldsymbol{\beta}^*||_1.$$

Then apply Theorem 3.3 to obtain that

$$
\begin{aligned}
||\widehat{\boldsymbol{\beta}}_{t+1}-\boldsymbol{\beta}^*||_2 \leqslant & \frac{a}{4\sqrt{s}}+\frac{(2b)}{4\sqrt{s}}||\widehat{\boldsymbol{\beta}}_t-\boldsymbol{\beta}^*||_1 \leqslant \frac{a}{4\sqrt{s}}+\frac{b}{4\sqrt{s}}\left(\frac{a(1-(2b)^t)}{1-(2b)}+(2b)^t||\widehat{\boldsymbol{\beta}}_0-\boldsymbol{\beta}^*||_1\right) \\
= & \frac{1}{4\sqrt{s}}\left(a+\frac{a((2b)-(2b)^{t+1})}{1-(2b)}\right)+\frac{(2b)^{t+1}||\widehat{\boldsymbol{\beta}}_0-\boldsymbol{\beta}^*||_1}{4\sqrt{s}} \\
= & \frac{a(1-(2b)^{t+1})}{4\sqrt{s}(1-(2b))}+\frac{(2b)^{t+1}||\widehat{\boldsymbol{\beta}}_0-\boldsymbol{\beta}^*||_1}{4\sqrt{s}},
\end{aligned}
$$

which concludes the proof. □

## A.5 Proof of Lemma 4.2

*Proof.* By the definition of $\mathcal{L}_j(\boldsymbol{\beta})$, we have

$$
\frac{1}{m}\sum_{j\in[m]}\nabla\mathcal{L}_j(\boldsymbol{\beta}^*)=\frac{1}{mn}\sum_{j\in[m]}\sum_{i\in[n]}\mathbf{x}_{ji}\epsilon_{ji}.
$$

Since $\epsilon_{ji}$ is mean zero subgaussian with $\phi_2$ norm bounded $\sigma$, then $\mathbf{x}_{ji}\epsilon_{ji}$ is also mean zero subgaussian with $\phi_2$ norm bounded $\sigma\max_{ji}(||\mathbf{x}_{ji}||_\infty)$, then apply Hoeffding-type inequality (Proposition 5.10 in (Vershynin, 2012)) and an union bound over $[p]$ leads to the desired bound. □

## A.6 Proof of Lemma 4.3

*Proof.* Applying Hoeffding-type inequality (Proposition 5.10 in (Vershynin, 2012)) and an union bound over $j\in[m], i\in[n]$ and $[p]$ leads to the desired bound. □

## A.7 Proof of Lemma 4.6

*Proof.* By the definition of $\mathcal{L}_j(\boldsymbol{\beta})$, we have

$$
\frac{1}{m}\sum_{j\in[m]}\nabla\mathcal{L}_j(\boldsymbol{\beta}^*)=\frac{1}{mn}\sum_{j\in[m]}\sum_{i\in[n]}\mathbf{x}_{ji}\left(y_{ji}-\frac{y_{ji}}{1+\exp(-y_{ji}\langle\boldsymbol{\beta},\mathbf{x}_{ji}\rangle)}\right).
$$

It is easy to check that

$$
\mathbb{E}\left[y_{ji}-\frac{y_{ji}}{1+\exp(-y_{ji}\langle\boldsymbol{\beta},\mathbf{x}_{ji}\rangle)}\right]=0,\quad\text{and}\quad\left|y_{ji}-\frac{y_{ji}}{1+\exp(--y_{ji}\langle\boldsymbol{\beta},\mathbf{x}_{ji}\rangle)}\right|\leqslant 1.
$$

and thus

$$
\begin{aligned}
&\mathbb{E}\left[\mathbf{x}_{ji}\left(y_{ji}-\frac{y_{ji}}{1+\exp(-y_{ji}\langle\boldsymbol{\beta},\mathbf{x}_{ji}\rangle)}\right)\right]=0,\\
&\left|\left|\mathbf{x}_{ji}\left(y_{ji}-\frac{y_{ji}}{1+\exp(--y_{ji}\langle\boldsymbol{\beta},\mathbf{x}_{ji}\rangle)}\right)\right|\right|_\infty\leqslant\max_{ji}\left(||\mathbf{x}_{ji}||_\infty\right).
\end{aligned}
$$

then apply Azuma-Hoeffding inequality (Hoeffding, 1963) and an union bound over $[p]$ leads to the desired bound. □